\documentclass[sigconf,screen,nonacm]{acmart}

\setcopyright{none}
\renewcommand\footnotetextcopyrightpermission[1]{}
\usepackage{cleveref}
\usepackage{algorithm}
\usepackage{algpseudocode}
\usepackage[table]{xcolor}
\usepackage{makecell}
\usepackage{pifont}
\usepackage{subcaption}
\usepackage{enumitem}
\AtBeginDocument{%
  }
    
\setcopyright{none}
\renewcommand\footnotetextcopyrightpermission[1]{}
\begin{document}

\title{LoMeVQA: A Comprehensive Benchmark for Longitudinal Medical VQA}
\author{Zhilin Wu}
\affiliation{%
  \institution{Tongji University}
  \city{Shanghai}
  \country{China}
}

\author{Zhangkai Ni}
\affiliation{%
  \institution{Tongji University}
  \city{Shanghai}
  \country{China}
}

\author{Chengmei Yang}
\affiliation{%
  \institution{Tongji University}
  \city{Shanghai}
  \country{China}
}

\author{Longzhen Yang}
\affiliation{%
  \institution{Tongji University}
  \city{Shanghai}
  \country{China}
}

\author{Yihang Liu}
\affiliation{%
  \institution{Tongji University}
  \city{Shanghai}
  \country{China}
}

\author{Ying Wen}
\affiliation{%
  \institution{East China Normal University}
  \city{Shanghai}
  \country{China}
}

\author{Lianghua He}
\affiliation{%
  \institution{Tongji University}
  \city{Shanghai}
  \country{China}
}

\begin{abstract}

In clinical practice, patients often undergo multiple imaging examinations over successive visits, yielding longitudinal data. Modeling such temporal information is crucial for reliable assessment of disease progression and treatment response.
However, despite the rapid advancement of multimodal large language models (MLLMs), longitudinal medical visual reasoning remains largely underexplored. 
To fill this gap, we propose \textbf{LoMeVQA}, a comprehensive benchmark consisting of 206K longitudinal visual question answering (VQA) pairs for temporal medical image analysis. LoMeVQA covers five tasks: progress classification, progress description, progress report generation, differential region grounding, and differential region description. 
To construct the dataset, we develop an automated pipeline that (1) organizes patient records chronologically, (2) extracts clinically meaningful entities via a medical knowledge graph, and (3) models their temporal evolution to guide large language models in generating high-quality longitudinal VQA pairs. 
Extensive evaluations demonstrate that both general-purpose and medical-domain MLLMs perform poorly on LoMeVQA, revealing substantial limitations in temporal reasoning. 
To address these limitations, we introduce \textbf{MedLong-8B}, which achieves state-of-the-art performance across all tasks. 
Beyond benchmarking, we conduct detailed analyses that uncover key failure modes and shed light on how to improve longitudinal medical visual reasoning. Our data is available  at: \url{https://github.com/pepperbubble/LoMeVQA}

\end{abstract}

\maketitle

\section{Introduction}
\begin{figure}[t]
    \centering
    \includegraphics[width=1\linewidth]{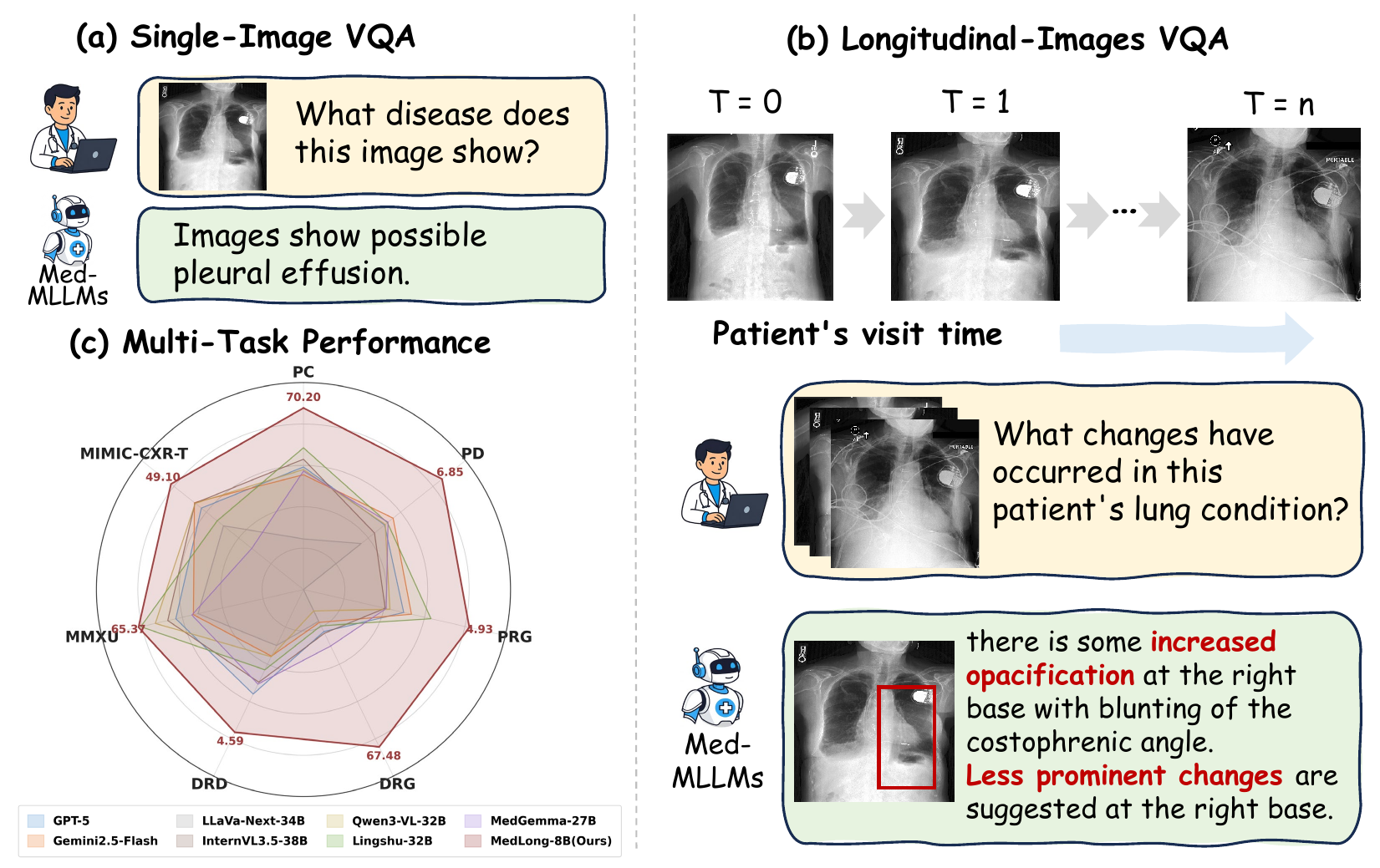}
    \caption{(a) Single-image analysis focuses on the current examination only.
    (b) Longitudinal image analysis leverages multiple examinations across visits to capture disease progression.
    (c) Performance comparison of different models. Task abbreviations are defined in Section~\ref{sec:taskoverview}.}
    \label{fig:teaser}
\end{figure}

\begin{figure*}[t]
    \centering
    \includegraphics[width=1\linewidth]{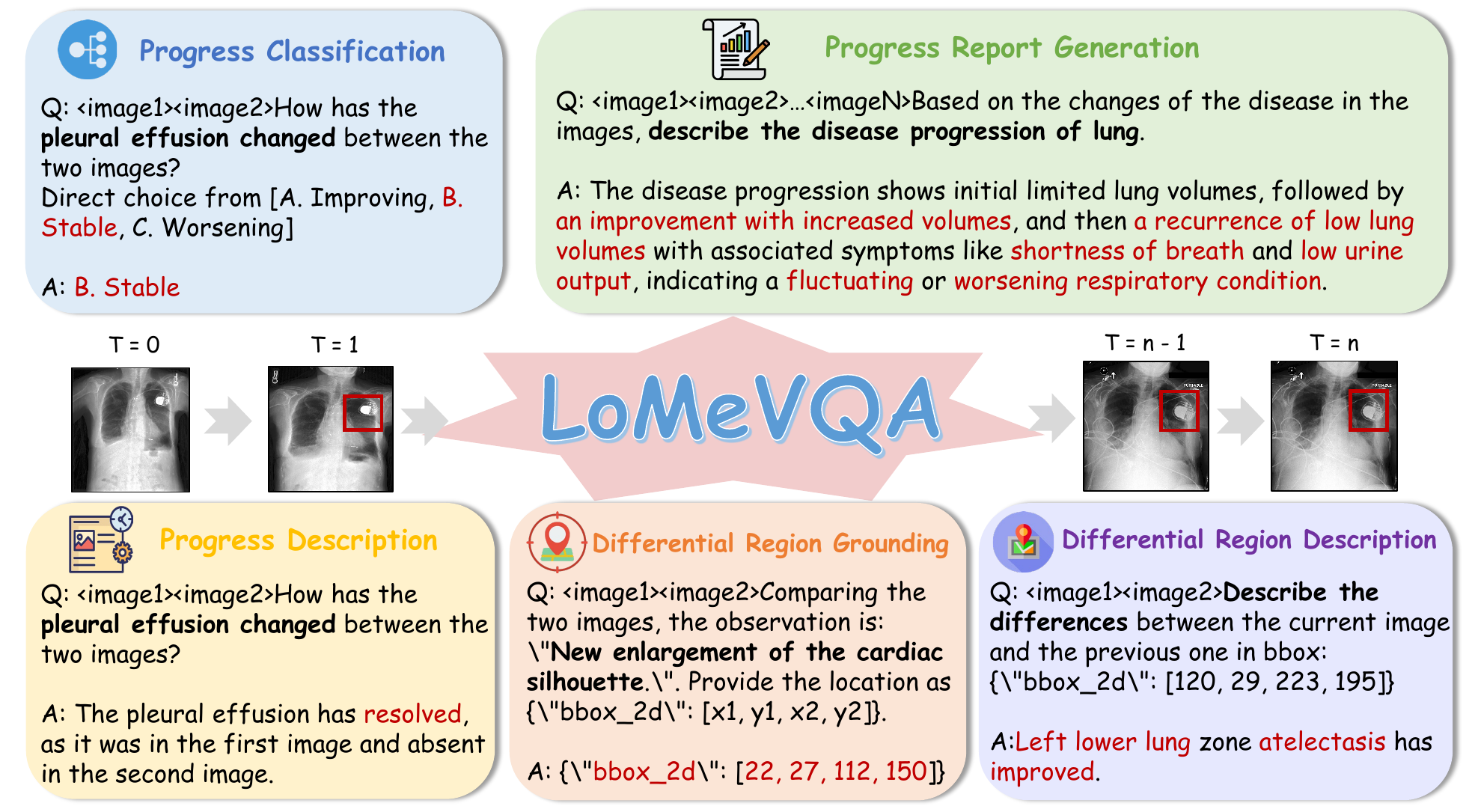}
    \caption{Representative examples from the five tasks in LoMeVQA, highlighting diverse scenarios of longitudinal medical visual reasoning. The benchmark supports comprehensive evaluation of multimodal large language models on longitudinal medical image understanding.}
    \label{fig:task_overview}
\end{figure*}
Multimodal large language models (MLLMs) have recently achieved remarkable progress in vision-language understanding and visual question answering across diverse general-domain benchmarks~\cite{hurst2024gpt,comanici2025gemini,bai2025qwen2,chen2024internvl,lu2024deepseek, alayrac2022flamingo}. 
Building upon these advances, medical adaptations of foundation models have shown substantial promise for medical image understanding and analysis~\cite{li2023llava, zambrano2025clinically, yang2025self, sellergren2025medgemma, ni2024m2trans}. 
Despite this progress, existing medical VQA and report generation datasets remain largely limited to single-visit scenarios, typically involving single-image inputs and short-form responses~\cite{lau2018dataset, he2020pathvqa, zhang2023pmc, bae2023ehrxqa}. This setting falls short of real-world clinical practice, where clinicians routinely compare longitudinal imaging studies and patient history to assess disease progression, treatment response, and prognosis.
As illustrated in~\Cref{fig:teaser}, longitudinal imaging consists of multiple examinations of the same patient acquired over time, forming a temporal sequence that captures disease evolution.
Interpreting such sequences is essential for identifying subtle anatomical and pathological changes and for supporting more accurate, personalized clinical decisions.
Although several recent studies have explored the use of longitudinal medical images~\cite{bannur2023learning, hu2023expert, mu2025mmxu, zhang2025libra, wang2024hergen, yang2024unlocking}, they are generally restricted to specific tasks and do not offer a systematic benchmark for evaluating diverse longitudinal reasoning abilities. As summarized in \Cref{tab:datasets_compare}, a comprehensive benchmark for longitudinal medical visual reasoning remains absent.

To address this gap, we introduce LoMeVQA,a comprehensive multi-task benchmark for longitudinal medical visual question answering, containing 206K VQA pairs. 
As illustrated in~\Cref{fig:task_overview}, LoMeVQA spans five representative tasks: progress classification, progress description, progress report generation, differential region grounding, and differential region description. 
Each sample contains two to five medical images acquired at different time points, together with a text question that examines temporal understanding and clinical reasoning. 
To construct LoMeVQA, we develop an automated pipeline that assembles longitudinal episodes according to visit chronology, extracts clinically meaningful entities using a medical knowledge graph, and models temporal evolution to guide large language models in synthesizing reliable question and answer pairs.
We benchmark both general-purpose and medical-domain MLLMs on LoMeVQA and observe substantial limitations in longitudinal reasoning. 
For instance, the best-performing models achieve only around 55\% accuracy on classification tasks and approximately 25\% performance on grounding tasks.
To improve performance in this setting, we develop \textbf{MedLong-8B} by fine-tuning Qwen3-VL-8B on the LoMeVQA-dev. 
MedLong-8B achieves state-of-the-art results across all tasks and provides deeper insights into the challenges of temporal medical image understanding. Our contributions can be summarized as follows:
\begin{itemize}
\item We introduce LoMeVQA, a large-scale multi-task benchmark for longitudinal medical image analysis. It covers classification, description, report generation, and spatial grounding tasks.

\item We develop an automated data construction pipeline for LoMeVQA. The pipeline integrates medical knowledge graph-based entity extraction and temporal evolution modeling to synthesize clinically grounded VQA pairs.

\item We systematically evaluate general and medical MLLMs on LoMeVQA. The results reveal substantial limitations in reasoning about temporal changes across visits.

\item We develop MedLong-8B, a fine-tuned variant of Qwen3-VL-8B trained on LoMeVQA. It achieves state-of-the-art performance across all tasks and establishes a strong baseline for longitudinal medical visual question answering.
\end{itemize}

\section{Related Works}
\begin{table*}[t]
  \centering
  \caption{Comparison of LoMeVQA with existing medical datasets. Longitudinal indicates whether the dataset contains temporal image sequences. Diversity indicates whether the answers are free-form rather than restricted to fixed templates. LoMeVQA is the first systematic benchmark for longitudinal medical image analysis in a multi-task setting. Medical-Diff-VQA* refers to the difference visual question answering subset within the Medical-Diff-VQA dataset.}
  \label{tab:datasets_compare}
  \begin{tabular}{l c c c c c c c c}
    \toprule
    dataset & year & size & Multi-Images & Longitudinal & Multi-Task & Diversity  & Tasks Format \\
    \midrule
    VQA-RAD & 2018 & 451 & \ding{55} &\ding{55} & \ding{55} & \ding{51} & MCQ \\
    Path-VQA & 2020 & 6,719  & \ding{55} &\ding{55} & \ding{55} & \ding{51} & MCQ \\
    MIMIC-CXR-T & 2023 & 1,326 & \ding{51} &\ding{51} & \ding{55} & \ding{55} & MCQ \\
    Medical-Diff-VQA* & 2023 & 164K & \ding{51} &\ding{51} & \ding{55} & \ding{55} & RG \\
    MMXU & 2025 & 118K & \ding{51} &\ding{51} & \ding{55} & \ding{51} & MCQ \\
    MeddSG-Bench & 2025 & 188K &\ding{51} & \ding{55} & \ding{55} & \ding{51} & Grounding \\
    \midrule
    \textbf{LoMeVQA} & \textbf{2026} & \textbf{206K} & \ding{51} & \ding{51} & \ding{51} & \ding{51} & \textbf{MCQ, RG, Grounding} \\
    \bottomrule
  \end{tabular}
\end{table*}

\textbf{Medical VQA Benchmarks}
The medical VQA task requires models to generate answers based on medical images and corresponding questions, playing an important role in assisting clinical diagnosis. This type of dataset usually contains multiple-choice questions or open-ended answers. Early work \cite{lau2018dataset,liu2021slake,he2020pathvqa, ben2019vqa, ben2021overview, zhang2023pmc, hu2024omnimedvqa} focused on single-image VQA tasks. Some recent studies \cite{zuo2025medxpertqa, yue2024mmmu, mu2025mmxu, yu2025medframeqa, yue2025medsg} have extended this field to multi-image VQA tasks. However, multi-images are not equal to longitudinal images. Longitudinal images refer to a series of scans from the same patient obtained at different time points during follow-up, thus capturing temporal changes in disease progression. In contrast, multi-image datasets typically comprise images collected across different patients, domains, or modalities. Longitudinal analysis of medical images provides critical insights into disease progression and patient prognosis. MIMIC-CXR-T \cite{bannur2023learning} proposed a longitudinal image classification task for the first time, providing labels for disease progression. Medical-Diff-VQA \cite{hu2023expert} introduced a longitudinal difference VQA dataset containing descriptions of disease changes. MMXU \cite{mu2025mmxu} increased sample diversity by incorporating anatomical location descriptions into the questions. Despite these contributions, these datasets either limit diversity or focus on a single type of task. To enable a comprehensive evaluation of longitudinal medical image analysis, we systematically construct a multi-task longitudinal image understanding dataset comprising 206K VQA samples and evaluate existing models on it.\\
\textbf{Multimodal Large Language Models in Medicine}
With the rapid development of general MLLMs \cite{liu2023visual, bai2025qwen2, chen2024internvl}, recent studies have begun adapting these models to the medical domain \cite{li2023llava, zambrano2025clinically,zhang2023huatuogpt, xu2025lingshu, sellergren2025medgemma}. LLaVA-Med\cite{li2023llava} used a two-stage instruction tuning to adapt LLaVA\cite{liu2023visual} for medical VQA tasks. MedGemma \cite{sellergren2025medgemma} combined SigLIP \cite{zhai2023sigmoid} and Gemma3 \cite{team2025gemma} to  perform medical multi-modal tasks. Lingshu\cite{xu2025lingshu} achieved the best performance in medical VQA and report generation by combining instruction tuning and reinforcement learning. Despite these achievements, most models are limited to single-image understanding, overlooking longitudinal medical analysis that requires fine-grained reasoning across multiple images. This paper aims to build a comprehensive longitudinal medical VQA dataset to facilitate the development of medical MLLMs.\\
\textbf{LLM-Generated Medical Datasets}
Large scale medical datasets are crucial for training MLLMs. Early medical VQA datasets were primarily constructed through manual annotation, resulting in data that were limited in scale. Recently, some studies expanded the scale by using LLMs to synthesize medical data\cite{yu2025medframeqa, huang2025m1, wu2025medreason, chen2024huatuogpt, xie2024medtrinity}. For example, Chen \cite{chen2024huatuogpt} used GPT-4o\cite{hurst2024gpt} to generate 40K medical complex reasoning data. However, directly using LLMs to generate medical data may produce hallucinations. To alleviate this problem, we leverage a knowledge graph \cite{jain2021radgraph} to guide data synthesis and improve dataset reliability.

\section{LoMeVQA Benchmark}
\begin{figure*}
    \centering
    \includegraphics[width=1\linewidth]{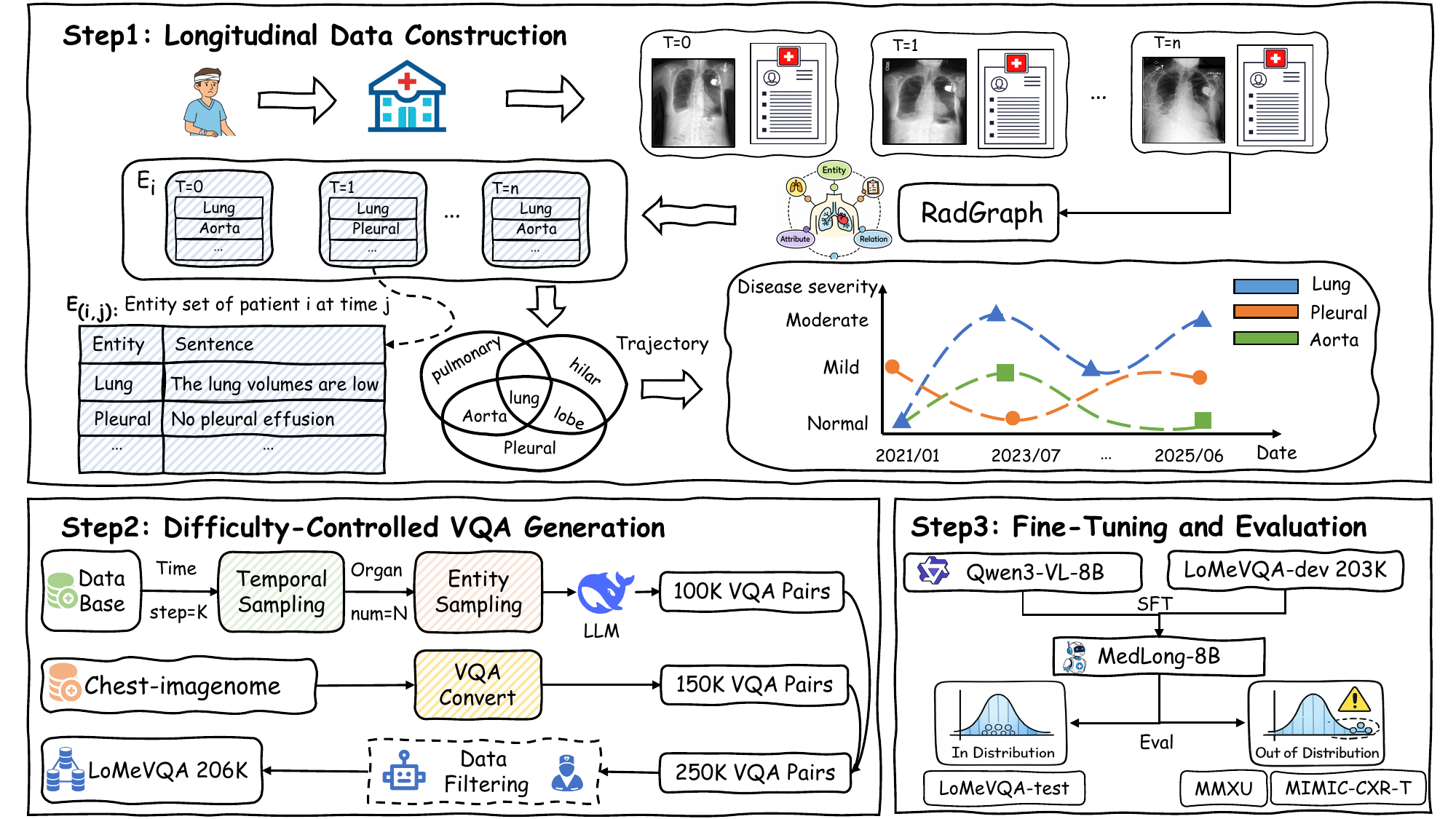}
    \caption{The pipeline of constructing LoMeVQA. Step(1) We collect longitudinal data from patients at different time points, where each sample includes an image and its corresponding report. We then extract entities and their descriptions using RadGraph. Step(2) We derive disease progression descriptions through temporal and entity sampling, and input them into LLMs to generate VQA pairs. Step(3) By fine-tuning Qwen3-VL-8B on LoMeVQA-dev, we obtain MedLong-8B, which is evaluated across multiple downstream tasks in both in-distribution and out-of-distribution settings.}
    \label{fig:pipeline}
\end{figure*}

In this section, we describe the task formulation and dataset construction process of LoMeVQA. 
As shown in~\Cref{fig:task_overview}, we define five visual question answering tasks for longitudinal medical image analysis: (1) Progress Classification; (2) Progress Description; (3) Progress Report Generation; (4) Differential Region Grounding; and (5) Differential Region Description.
The overall dataset construction includes two main components: longitudinal data construction and difficulty-controlled VQA synthesis. 
We integrate medical knowledge graphs with LLMs to automate the generation process. 
After quality control and filtering, the final LoMeVQA dataset contains 206K VQA samples in total. To construct a reliable evaluation benchmark, we further perform expert review and select 2,500 high-quality samples as the LoMeVQA-test set, while the remaining samples are used to form LoMeVQA-dev. Further statistical information for each task is presented in \Cref{tab:dataset_statistic} in Appendix~\labelcref{sec:dataset_statistic}.

\subsection{Task Overview}
\label{sec:taskoverview}

\textbf{Task 1: Progress Classification (PC)}.  
This task takes as input two longitudinal medical images from the same patient along with a disease-specific question, and requires the model to classify the disease progression as improved, stable, or worsened. 
Clinically, this corresponds to the evaluation process in which radiologists compare a follow-up examination to a previous one to determine the direction of disease evolution.
Example: Given two chest radiographs and the question ``Has the pulmonary opacity improved?”, the correct answer may be ``Worsened”, indicating an increase in the extent or density of the opacity over time.

\textbf{Task 2: Progress Description (PD)}.  
This task requires the model to analyze two longitudinal medical images and generate a descriptive explanation of how the relevant disease findings have changed. 
This aligns with the descriptive interpretation commonly included in follow-up radiology reports. 
The task promotes interpretability by requiring explicit articulation of observed progression patterns.
Example: Given two chest radiographs and the question ``How has the cardiac enlargement changed between the two images?”, an appropriate answer is ``The cardiac enlargement has increased from mildly enlarged in the first image to more clearly enlarged in the second image.”

\textbf{Task 3: Progress Report Generation (PRG)}.  
This task involves synthesizing multiple longitudinal medical images together with questions concerning several anatomical regions, intending to generate an integrated report that summarizes disease progression across time. 
Clinically, this reflects the workflow in which radiologists incorporate findings across organs and visits to construct a cohesive longitudinal assessment.
Example: Given a series of chest radiographs and the question ``Based on the changes across the images, describe the progression of pleural findings.”, the answer may narrate the emergence, development, and resolution of pleural effusion. 
Unlike Task 2, which focuses on a single lesion across two time points, Task 3 synthesizes observations across multiple time points and regions.

\textbf{Task 4: Differential Region Grounding (DRG)}.
This task requires the model to identify and localize the regions that exhibit notable changes between two longitudinal medical images from the same patient, in response to a fixed question such as ``Where has the current image changed compared to the previous image?”.
Clinically, accurate localization of evolving lesions is essential for treatment monitoring and therapeutic planning. 
Visual grounding also provides interpretability, allowing clinicians to verify model predictions.
Example: Given two chest radiographs and the question, the answer may be a bounding box such as ``{[31, 48, 109, 176]}”.

\textbf{Task 5: Differential Region Description (DRD)}.  
This task requires the model to describe the specific difference within a predefined region between two longitudinal medical images. 
Clinicians often evaluate disease progression in localized regions to support targeted clinical decision making.
Example: Given two chest radiographs and the question ``Describe the differences in the region {[7, 18, 104, 131]}.”, the answer may be ``There is resolution of the previously observed large right-sided pleural effusion.”

\subsection{Dataset Construction}
\textbf{Longitudinal data construction} 
As shown in~\Cref{fig:pipeline}, we construct patient-level longitudinal episodes from the original MIMIC CXR dataset~\cite{johnson2019mimic}, so that each sample reflects the temporal progression of a patient's condition. As shown in \Cref{alg:dataset}, we denote the dataset as $\mathcal{D} = \{ P_1, P_2, \dots, P_d \}$, where $d$ is the total number of patients. 
Each patient $P_i$ has a time-ordered study sequence:
\begin{equation}
P_i = ( S_{(i,1)}, S_{(i,2)}, \dots, S_{(i,n_i)} ), \quad i = 1, \dots, d,    
\end{equation}
where $n_i$ is the number of studies for patient $P_i$. 
Each study $S_{(i,j)}$ contains an image and report pair:
\begin{equation}
    S_{(i,j)} = (I_{(i,j)}, R_{(i,j)}), \quad j = 1, \dots, n_i.
\end{equation}

Each report $R_{(i,j)}$ records diagnostic findings and clinical observations at the corresponding time point. 
We employ RadGraph~\cite{jain2021radgraph} to extract structured medical entities and their supporting sentence-level descriptions:
\begin{equation}
E_{(i,j)}=\mathrm{RadGraph}(R_{(i,j)}).
\end{equation}

For each entity $e$ in $E_{(i,j)}$, we retrieve the corresponding sentence $x_{(i,j)}^e\in R_{(i,j)}$. 
For example, for the $j$-th visit of patient $P_i$, we obtain an entity set $E_{(i,j)}$, which may include entities such as ``lung'' with its corresponding text “The lung volumes are low.” 
Some entities may appear in reports across multiple time points. 
To capture their temporal evolution, we aggregate the descriptions of the same entity across all visits of a patient and construct an entity-level trajectory ${T_i^e}$, which reflects the longitudinal changes of the entity $e$ over time. 
Finally, we obtain the entity set $E$ and the corresponding trajectory set $\mathcal{T}=\{T_i^e\}$.

\begin{algorithm}[ht]
\caption{LoMeVQA Dataset Construction Pipeline}
\label{alg:dataset}
\begin{algorithmic}[1]
\Require Original dataset $\mathcal{D}=\{P_1, P_2, \dots, P_d\}$, sampled entity number $N$, window length $K$, predefined prompt $\mathrm{Prompt}$, large language model $\mathrm{LLM}$
\Ensure Final dataset $\mathcal{D}_{\mathrm{LoMeVQA}} = \{(V_m, q_m, a_m)\}_{m=1}^{|\mathcal{D}_{\mathrm{LoMeVQA}}|}$, where each $(V_m,q_m,a_m)$ is a VQA sample

\State \textbf{Initialize:} $\mathcal{D}_{\mathrm{LoMeVQA}} \gets \emptyset$, $E \gets \emptyset$, $\mathcal{T} \gets \emptyset$

\State \textcolor{blue}{$\triangleright$ Step 1: Longitudinal data construction}
\For{each patient $P_i \in \mathcal{D}$}
    \State $E_i \gets \emptyset$
    \For{each study $S_{(i,j)}=(I_{(i,j)},R_{(i,j)})$ in $P_i$}
        \State $E_{(i,j)} \gets \mathrm{RadGraph}(R_{(i,j)})$
        \State $E_i \gets E_i \cup E_{(i,j)}$
        \For{each entity $e \in E_{(i,j)}$}
            \State $x_{(i,j)}^e \gets \mathrm{GetSentence}(R_{(i,j)}, e)$
            \State Append $x_{(i,j)}^e$ to $T_i^e$
        \EndFor
    \EndFor
    \State $E \gets E \cup E_i$
    \State $\mathcal{T} \gets \mathcal{T} \cup \{T_i^e \mid e \in E_i\}$
\EndFor

\State \textcolor{blue}{$\triangleright$ Step 2: Difficulty-controlled VQA generation}
\For{each patient $P_i \in \mathcal{D}$}
    \State Extract the ordered studies $P_i = (S_{(i,1)}, S_{(i,2)}, \dots, S_{(i,n_i)})$
    \State \textcolor{blue}{$\triangleright$ temporal sampling}
    \For{$j = 1$ to $n_i-K+1$}
        \State Form a study window $C = (S_{(i,j)}, \dots, S_{(i,j+K-1)})$
        \State Collect the entity sets $\{E_{(i,j)}, E_{(i,j+1)}, \dots, E_{(i,j+K-1)}\}$
        \State Compute the shared entity set $E^{*}=\bigcap_{k=0}^{K-1} E_{(i,j+k)}$
        \If{$|E^{*}| \ge N$}
            \State \textcolor{blue}{$\triangleright$ entity sampling}
            \State Randomly sample $O \subseteq E^{*}$ such that $|O| = N$
            \State Get $X \gets \{T_i^e \mid e \in O\}$
            \State Generate a pair $(q,a) \gets \mathrm{LLM}(\mathrm{Prompt}, X)$
            \State Collect the corresponding image sequence $V$ 
            \State $\mathcal{D}_{\mathrm{LoMeVQA}} \gets \mathcal{D}_{\mathrm{LoMeVQA}} \cup \{(V,q,a)\}$
        \EndIf
    \EndFor
\EndFor

\end{algorithmic}
\end{algorithm}

\textbf{Difficulty-Controlled VQA generation}
Based on the constructed longitudinal data, we further generate VQA samples with controllable difficulty. To achieve this, we design the question generation process through temporal sampling and entity sampling.
For patient $P_i$, we use a sliding window $C$ of length $K$ to obtain a subsequence, where $C$ represents a consecutive series of studies and satisfies $|C| = K$:
\begin{equation}
C = ( S_{(i,j)}, S_{(i,(j+1))}, \dots, S_{(i,(j+K-1))} ).    
\end{equation}

For each subsequence $C$, we collect the corresponding entity sets$\{ E_{(i,j)}, E_{(i,(j+1))}, \dots, E_{(i,(j+K-1))} \}$ associated with the studies in $C$. Their intersection is computed to identify the shared clinical entities within this window:
\begin{equation}
    E^{*} = \bigcap_{k=0}^{K-1} E_{(i,j+k)}, \quad |E^{*}| = M.
\end{equation}

The intersection set $E^*$ contains $M$ shared entities. 
We randomly sample $N$ entities from this set:
\begin{equation}
    {O} \subseteq {E^*}, \quad |{O}| = N.
\end{equation}

Finally, for each patient $P_i$, we obtain a set of entities derived from the intersection process:
\begin{equation}
    {O} = \{ e_1, e_2, \dots, e_N \}.
\end{equation}

For each entity $e \in O$, we extract its corresponding trajectory ${T}_i^{e}$, thereby forming a descriptive sentence collection $X$ that captures the longitudinal changes of these entities across visits. 
We then provide $X$ together with a predefined prompt to a large language model to generate a question–answer pair:
\begin{equation}
    (q, a) = LLM(Prompt, X).
\end{equation}

Together with the corresponding image sequence
\begin{equation}
    V=(I_{(i,j)},I_{(i,(j+1))},\dots,I_{(i,(j+K-1))}),
\end{equation}
we obtain a final sample $(V, q, a)$.
\Cref{fig:case_generation} in Appendix~\ref{sec:vqa_generation} provides an example of VQA generation.

Overall, this process enables the synthesis of VQA samples with controllable difficulty by controlling the temporal span and the number of involved entities. In addition, by incorporating the corresponding descriptions from the original clinical reports during question generation, we substantially reduce hallucinations in the generated content. We construct the first three tasks (Progress Classification, Progress Description, and Progress Report Generation) following the above automated pipeline. For the remaining two tasks (Differential Region Grounding and Differential Region Description) that require region-level annotations, we convert existing datasets \cite{wu2021chest} into the VQA format using predefined question templates. Together, these components form the comprehensive LoMeVQA benchmark for longitudinal medical image analysis.

\begin{table}[t]
  \centering
  \caption{Consistency filter using LLM.}
  \label{tab:consistency_filter}
  \begin{tabular}{lccc}
    \toprule
    Task & Consistent & Contradictory & Uncertain \\
    \midrule
    PC & 73.8\% & 22.1\% & 4.1\% \\
    PD & 90.1\% & 6.1\% & 3.8\% \\
    PRG & 94.6\% & 4.8\% & 0.6\% \\
    \bottomrule
  \end{tabular}
\end{table}

\textbf{Data Filtering} 
Ensuring the reliability of data synthesized by large language models remains a central challenge. 
Although the proposed pipeline reduces hallucination by anchoring question generation to sentence-level descriptions extracted from original clinical reports, the generated VQA samples may still exhibit semantic inconsistency or imbalanced answer distributions. 
To address these issues, we apply a two-stage filtering procedure focusing on diversity and consistency.

For \emph{diversity filtering}, we analyze the distribution of answers within each task and remove subsets whose responses display excessive concentration beyond a predefined threshold. 
This step encourages broader coverage of clinically meaningful variations and prevents over-representation of trivial or default answers.

For \emph{consistency filtering}, we design an automatic verification pipeline to evaluate factual alignment. 
Specifically, for each generated sample $(V,q,a)$, we retrieve the source clinical report and prompt an LLM to assess the relationship between $(q,a)$ and the report content, labeling it as consistent, contradictory, or uncertain. 
Only samples classified as consistent are retained. 
This procedure ensures that the resulting dataset maintains coherence with real clinical evidence and reduces propagation of erroneous interpretations. As shown in~\Cref{tab:consistency_filter}, Tasks 2 and 3 achieve consistency pass rates above 90\%, whereas Task 1 yields a lower rate of 73\%. 
The lower performance in Task 1 is attributable to its three-option multiple-choice setting, which increases the probability of generating semantically conflicting responses. 
A representative contradictory case is illustrated in \Cref{fig:contradictory} in Appendix~\labelcref{sec:data_filter}. 

After the initial LLM-based filtering, we further conduct a human review stage to construct a reliable test benchmark. Three clinicians with different levels of expertise assess the retained samples. We provide the corresponding human review pass rates on LoMeVQA-test in \Cref{tab:human_check} in Appendix~\ref{sec:data_filter}. In the final benchmark construction, we retain only the 2,500 samples that are approved by both the automatic filtering pipeline and human reviewers, ensuring the high quality and clinical reliability of the test set.

\begin{figure}
    \centering
    \includegraphics[width=1\linewidth]{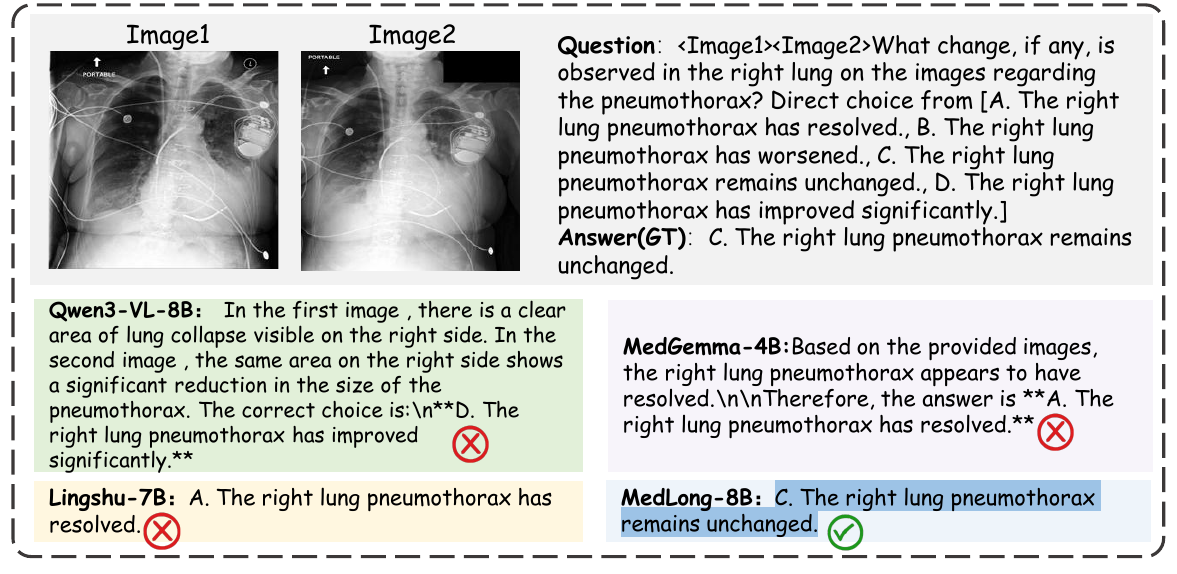}
    \caption{MedLong-8B shows generalization ability on the MMXU-test, even though it is not trained on it.}
    \label{fig:case_study_mmxu}
\end{figure}

\definecolor{green-1}{rgb}{0.388,0.745,0.482}
\definecolor{green-2}{rgb}{0.627,0.847,0.686}
\definecolor{green-3}{rgb}{0.871,0.949,0.890}
\definecolor{lightgrayrow}{gray}{0.93}

\newcommand{\best}[2]{%
  \begingroup
  \setlength{\fboxsep}{0.6pt}%
  \colorbox{#1}{\strut #2}%
  \endgroup
}
\newcommand{\triple}[3]{%
  \makebox[10em][c]{#1 / #2 / #3}%
}
\newcommand{\groupheader}[1]{%
  \rowcolor{lightgrayrow}
  \multicolumn{7}{c}{\textbf{#1}} \\
}

\begin{table*}[t]
  \centering
  \caption{Performance of different MLLMs on LoMeVQA-test. Task names and metrics are abbreviated. Acc: Accuracy; CE: Clinical Efficacy; NLG: Natural Language Generation. We calculate average metrics on LoMeVQA-test. N/A means that the input length exceeds the maximum length of the model.}
  \label{tab:performance}
  \setlength{\tabcolsep}{1pt}
  \begin{tabular}{lcccccc}
    \toprule
     & PC & PD & PRG & DRG & DRD & Average \\
    \midrule
    Metric & Acc & CE / NLG / LLM & CE / NLG / LLM & IoU & CE / NLG / LLM & \\
    \midrule

    \groupheader{General MLLMs}
    GPT-4o & 38.40
      & \makecell[c]{\triple{14.36}{20.84}{3.99}}
      & \makecell[c]{\triple{10.79}{20.61}{2.95}}
      & 16.91
      & \makecell[c]{\triple{2.83}{11.24}{3.43}}
      & 17.13 \\
    GPT-5 & 47.40
      & \makecell[c]{\triple{18.30}{26.92}{4.01}}
      & \makecell[c]{\triple{11.77}{21.90}{2.98}}
      & 18.06
      & \makecell[c]{\triple{4.83}{12.99}{3.36}}
      & 20.23 \\
    Gemini2.5-Flash & 44.40
      & \makecell[c]{\triple{23.55}{28.66}{4.43}}
      & \makecell[c]{\triple{14.28}{22.83}{3.21}}
      & 14.11
      & \makecell[c]{\triple{2.44}{11.42}{2.14}}
      & 19.23 \\
    LLaVa-Next-7B & 19.60
      & \makecell[c]{\triple{11.81}{14.15}{3.07}}
      & \makecell[c]{N/A}
      & 9.34
      & \makecell[c]{\triple{0.72}{6.61}{1.88}}
      & 8.34 \\
    LLaVa-Next-13B & 13.20
      & \makecell[c]{\triple{11.88}{14.54}{3.04}}
      & \makecell[c]{N/A}
      & 10.64
      & \makecell[c]{\triple{0.77}{6.83}{1.98}}
      & 7.37 \\
    LLaVa-Next-34B & 19.60
      & \makecell[c]{\triple{15.56}{16.49}{2.85}}
      & \makecell[c]{N/A}
      & 13.61
      & \makecell[c]{\triple{0.69}{6.85}{1.80}}
      & 9.59 \\
    InternVL3.5-8B & 32.40
      & \makecell[c]{\triple{19.40}{25.07}{3.16}}
      & \makecell[c]{\triple{8.43}{18.39}{2.49}}
      & 8.83
      & \makecell[c]{\triple{1.93}{6.83}{2.56}}
      & 14.13 \\
    InternVL3.5-14B & 46.00
      & \makecell[c]{\triple{20.68}{26.58}{3.44}}
      & \makecell[c]{\triple{7.98}{17.44}{2.44}}
      & 12.22
      & \makecell[c]{\triple{2.53}{7.32}{2.38}}
      & 17.70 \\
    InternVL3.5-38B & 50.40
      & \makecell[c]{\triple{22.28}{27.82}{3.51}}
      & \makecell[c]{\triple{8.53}{19.07}{2.43}}
      & 18.72
      & \makecell[c]{\triple{3.62}{10.13}{2.98}}
      & 20.52 \\
    Qwen3-VL-4B & 34.80
      & \makecell[c]{\triple{10.43}{17.70}{3.94}}
      & \makecell[c]{\triple{9.91}{15.59}{3.85}}
      & 19.96
      & \makecell[c]{\triple{1.42}{7.36}{1.88}}
      & 15.76 \\
    Qwen3-VL-8B & 33.80
      & \makecell[c]{\triple{9.43}{15.74}{3.64}}
      & \makecell[c]{\triple{8.16}{15.56}{3.18}}
      & 6.83
      & \makecell[c]{\triple{1.52}{7.24}{2.03}}
      & 12.56 \\
    Qwen3-VL-32B & 46.40
      & \makecell[c]{\triple{8.41}{14.54}{4.04}}
      & \makecell[c]{\triple{7.02}{14.19}{2.57}}
      & 9.14
      & \makecell[c]{\triple{1.98}{5.30}{2.16}}
      & 15.12 \\
    \midrule

    \groupheader{Medical MLLMs}
    Lingshu-7B & 52.00
      & \makecell[c]{\triple{19.78}{26.24}{3.87}}
      & \makecell[c]{\triple{13.12}{20.33}{2.59}}
      & 8.07
      & \makecell[c]{\triple{3.87}{12.56}{2.61}}
      & 19.01 \\
    Lingshu-32B & 54.80
      & \makecell[c]{\triple{26.67}{29.77}{4.15}}
      & \makecell[c]{\triple{16.45}{17.67}{3.79}}
      & 15.54
      & \makecell[c]{\triple{3.32}{12.60}{2.58}}
      & 21.87 \\
    MedGemma-4B & 52.60
      & \makecell[c]{\triple{20.95}{20.28}{4.74}}
      & \makecell[c]{\triple{10.73}{16.26}{3.48}}
      & 12.07
      & \makecell[c]{\triple{4.34}{8.90}{2.96}}
      & 19.11 \\
    MedGemma-27B & 45.80
      & \makecell[c]{\triple{12.93}{17.79}{4.18}}
      & \makecell[c]{\triple{7.34}{14.69}{2.46}}
      & 24.21
      & \makecell[c]{\triple{2.94}{7.40}{3.04}}
      & 18.85 \\
    \midrule

    \groupheader{Fine-tuned MLLMs}
    MedLong-4B(single) & \cellcolor{green-3}69.76
      & \makecell[c]{\triple{40.02}{43.61}{6.33}}
      & \makecell[c]{\triple{32.33}{32.64}{4.37}}
      & 64.71
      & \makecell[c]{\triple{\best{green-1}{10.50}}{16.02}{4.16}}
      & 39.56 \\
    MedLong-4B(full) & \cellcolor{green-2}69.84
      & \makecell[c]{\triple{\best{green-2}{42.34}}{\best{green-2}{46.14}}{\best{green-3}{6.71}}}
      & \makecell[c]{\triple{\best{green-3}{32.70}}{\best{green-3}{33.75}}{\best{green-3}{4.64}}}
      & \cellcolor{green-3}65.98
      & \makecell[c]{\triple{\best{green-3}{10.27}}{\best{green-2}{17.17}}{\best{green-3}{4.21}}}
      & \best{green-3}{40.36} \\
    MedLong-8B(single) & 68.93
      & \makecell[c]{\triple{\best{green-3}{42.24}}{\best{green-3}{46.03}}{\best{green-2}{6.74}}}
      & \makecell[c]{\triple{\best{green-1}{33.40}}{\best{green-1}{35.22}}{\best{green-2}{4.89}}}
      & \cellcolor{green-1}67.64
      & \makecell[c]{\triple{10.09}{\best{green-1}{17.72}}{\best{green-1}{4.65}}}
      & \best{green-2}{40.71} \\
    MedLong-8B(full) & \cellcolor{green-1}70.20
      & \makecell[c]{\triple{\best{green-1}{45.32}}{\best{green-1}{47.83}}{\best{green-1}{6.85}}}
      & \makecell[c]{\triple{\best{green-2}{33.13}}{\best{green-2}{33.97}}{\best{green-1}{4.93}}}
      & \cellcolor{green-2}67.48
      & \makecell[c]{\triple{\best{green-2}{10.31}}{\best{green-3}{16.75}}{\best{green-2}{4.59}}}
      & \best{green-1}{41.14} \\
    \bottomrule
  \end{tabular}
\end{table*}
\section{Experiments}
\begin{table}[t]
  \centering
  \setlength{\tabcolsep}{3pt}
  \caption{Performance on the OOD benchmarks MMXU-test and MIMIC-CXR-T. The green upward arrow denotes the absolute gain of MedLong-8B over Qwen3-VL-8B.}
  \label{tab:OOD}
  \begin{tabular}{lccc}
    \toprule
    \textbf{Model} & \textbf{MMXU-test} & \textbf{MIMIC-CXR-T} & \textbf{Average} \\
    \midrule
    GPT-5 & 50.60 & 37.86 & 44.23 \\
    Gemini2.5-Flash & 43.61 & 40.42 & 42.02 \\
    LLaVa-Next-34B & 41.80 & 29.64 & 35.72 \\
    InternVL3.5-38B & 53.80 & 40.20 & 47.00 \\
    Lingshu-32B & 64.47 & 31.98 & 48.23 \\
    MedGemma-27B & 44.23 & 19.16 & 31.70 \\
    Qwen3-VL-32B & 58.73 & 40.12 & 49.43 \\
    \midrule
    Qwen3-VL-8B & 45.03 & 38.69 & 41.86 \\
    \textbf{MedLong-8B} & \textbf{65.37} {\color{green-1}$(\uparrow$20.34)} & \textbf{49.10} {\color{green-1}$(\uparrow$10.41)} & \textbf{57.24} {\color{green-1}$(\uparrow$15.38)} \\
    \bottomrule
  \end{tabular}
\end{table}
\label{sec:experiments}
\subsection{Main Settings}

\textbf{Implementation Details}. 
All general and medical MLLMs are evaluated on the LoMeVQA-test. 
To further enhance longitudinal reasoning capability, we develop MedLong-4B and MedLong-8B by fine-tuning Qwen3-VL models on LoMeVQA-dev. 
During supervised fine-tuning, we adopt LoRA~\cite{hu2022lora} for parameter-efficient training, with a batch size of 64 and a learning rate of 2e-5. 
All models are trained for one epoch using the LLaMA-Factory framework\cite{zheng2024llamafactory}.

\textbf{Metric}. We use accuracy to evaluate performance on multiple-choice tasks. 
For open-ended generation tasks, we adopt a complementary set of evaluation metrics to assess response quality from different perspectives. Specifically, we report ROUGE~\cite{lin2004rouge} and METEOR~\cite{banerjee2005meteor} as natural language generation (NLG) metrics to measure textual similarity and overall generation quality, F1-RadGraph~\cite{delbrouck2022improving} as a clinical efficacy (CE) metric to evaluate the correctness of clinically meaningful entities and relations, and LLM-as-a-Judge as an additional metric to provide a more holistic assessment of medical accuracy, logical consistency, and clinical relevance. We provide the prompt for LLM-as-a-Judge in Appendix~\ref{llm-as-a-judge}. For grounding tasks, we use the Intersection over Union (IoU) score to assess localization performance.


\subsection{Main Results}
Based on the extensive experimental results, we summarize the main findings from \Cref{tab:performance} as follows.

\textbf{Longitudinal medical image analysis is still a challenge in MLLMs}.
The evaluation results on LoMeVQA indicate that existing MLLMs perform poorly across many tasks, and in some cases are unable to handle them effectively. 
For example, in the Progress Classification task, the best-performing medical model Lingshu achieves only around 55\% accuracy. 
In the Differential Region Grounding task, the highest IoU score among evaluated models is merely 25\%, while several models fail to produce valid grounding outputs. 
These results indicate that current models struggle with temporal comparison, subtle change detection, and spatial attribution over serial examinations.

\textbf{Medical models outperform general models}.
Lingshu \cite{xu2025lingshu} and MedGemma \cite{sellergren2025medgemma} are two state-of-the-art medical MLLMs that have been extensively pre-trained on large-scale medical datasets. 
Experiments in \Cref{tab:performance} show that both medical models exhibit better performance than general-purpose models on LoMeVQA. 
For instance, Lingshu-32B achieves notable improvements in progress classification, progress report generation, and differential region description, reflecting its enhanced ability to integrate temporal imaging cues with medical semantics. 
However, its performance in the differential region grounding task remains limited, indicating that precise spatial alignment of temporal changes remains difficult even for medically specialized models.

\textbf{Larger Models Achieve Better Fine-tuning Results}. To verify the fine-tuning effect of different parameter models, we develop MedLong-4B and MedLong-8B. \Cref{tab:performance} shows that MedLong-8B outperforms MedLong-4B most of the time. This indicates that increasing model capacity improves adaptation to longitudinal medical understanding tasks.

\begin{figure*}
    \centering
    \includegraphics[width=1\linewidth]{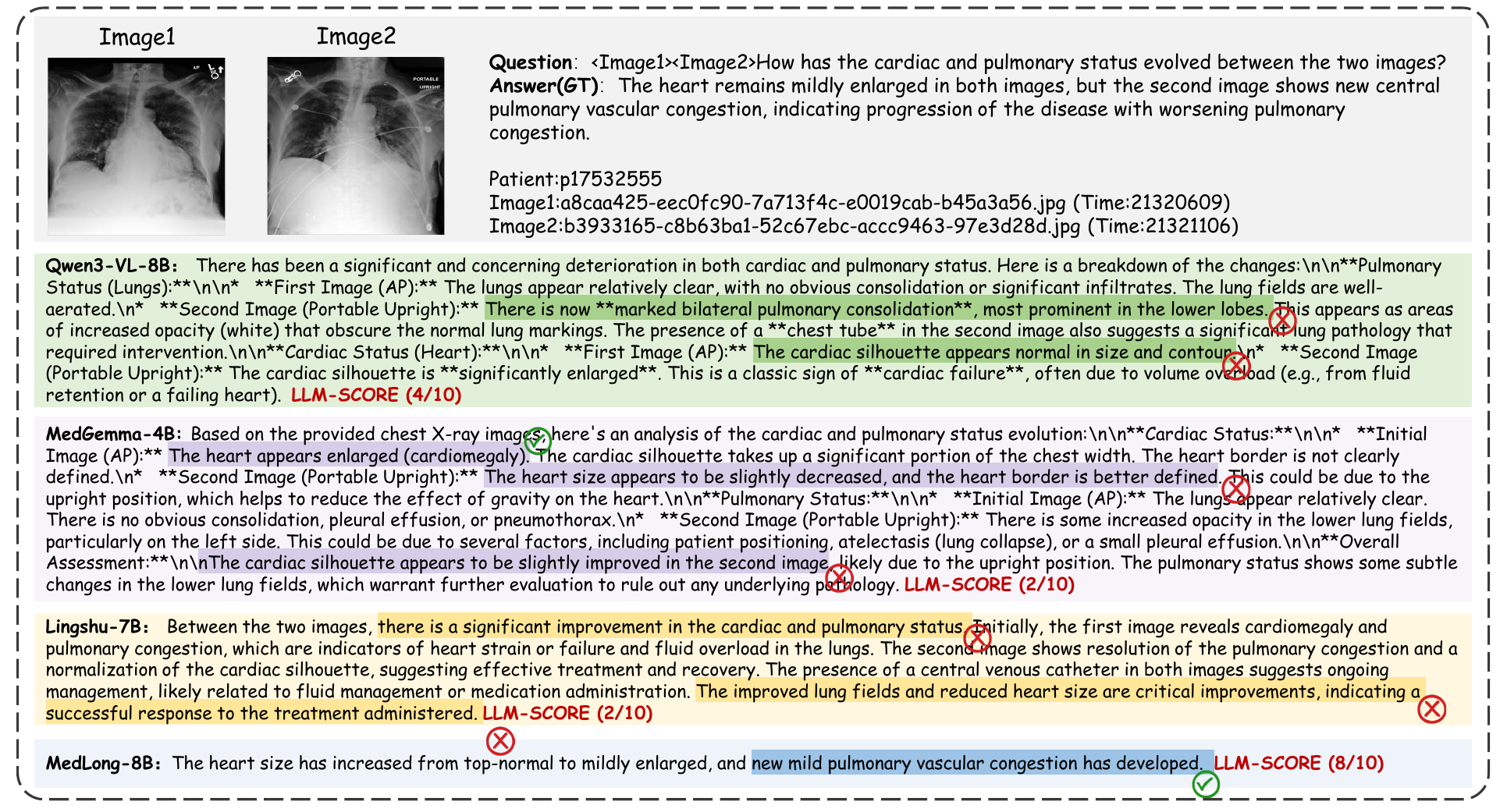}
    \caption{A case study for progress description. MedLong can capture and describe the changes between images.}
    \label{fig:case_study2}
\end{figure*}

\textbf{Does single-task fine-tuning outperform multi-task fine-tuning?} 
There are two approaches in the fine-tuning process: one is to train a model on each task separately, and the other is to train a model on all tasks together. Some previous work \cite{liu2024moe, crawshaw2020multi, li2020dice} claims that multi-task fine-tuning may hurt performance due to issues such as data imbalance and seesaw effects. 
But in our experiments, we observe that the performance of models fine-tuned on all tasks jointly (MedLong-8B-full) is comparable to that of models fine-tuned on individual tasks (MedLong-8B-single), suggesting that a sufficiently large model can learn multiple task-specific patterns simultaneously without substantial interference.

\textbf{Can MedLong-8B generalize to other longitudinal datasets?} 
We further evaluate different models on two additional longitudinal benchmarks, MMXU-test and MIMIC-CXR-T, to assess out-of-distribution generalization. MMXU-test contains 3,000 chest X-ray VQA samples and focuses on region-level disease progression. MIMIC-CXR-T contains 1,326 samples and evaluates progression-oriented question answering over five thoracic diseases. Both datasets involve longitudinal chest X-ray understanding, while their text annotations and task formats differ from LoMeVQA. As shown in \Cref{tab:OOD}, MedLong-8B demonstrates strong generalization ability on both benchmarks, even though it is not trained on either dataset. A qualitative example on MMXU-test is provided in \Cref{fig:case_study_mmxu}.

\subsection{Case Study and Qualitative Analysis}
In this section, we present representative case studies along with qualitative analyses to better interpret model behaviors. Additional examples are provided in the Appendix~\labelcref{sec:case_studies}.

\textbf{Case Study}. \Cref{fig:case_study2} illustrates an example of the PD task. MedLong-8B correctly identifies the longitudinal worsening pattern and produces a clinically coherent description consistent with the radiological progression. In contrast, existing models reveal clear limitations in longitudinal visual reasoning. General-domain models tend to over-rely on textual priors while overlooking image evidence. For instance, Qwen3-VL-8B incorrectly interprets the second image by identifying a normal lung region as pulmonary consolidation, and this perception error propagates to the subsequent reasoning process. Medical MLLMs such as Lingshu and MedGemma suffer less from severe misperception, yet they still fail to recognize the fine-grained worsening pulmonary congestion in the image sequence. This case highlights that accurate longitudinal medical understanding requires not only domain knowledge, but also reliable fine-grained visual perception across time.

\begin{figure}
    \centering
    \includegraphics[width=1\linewidth]{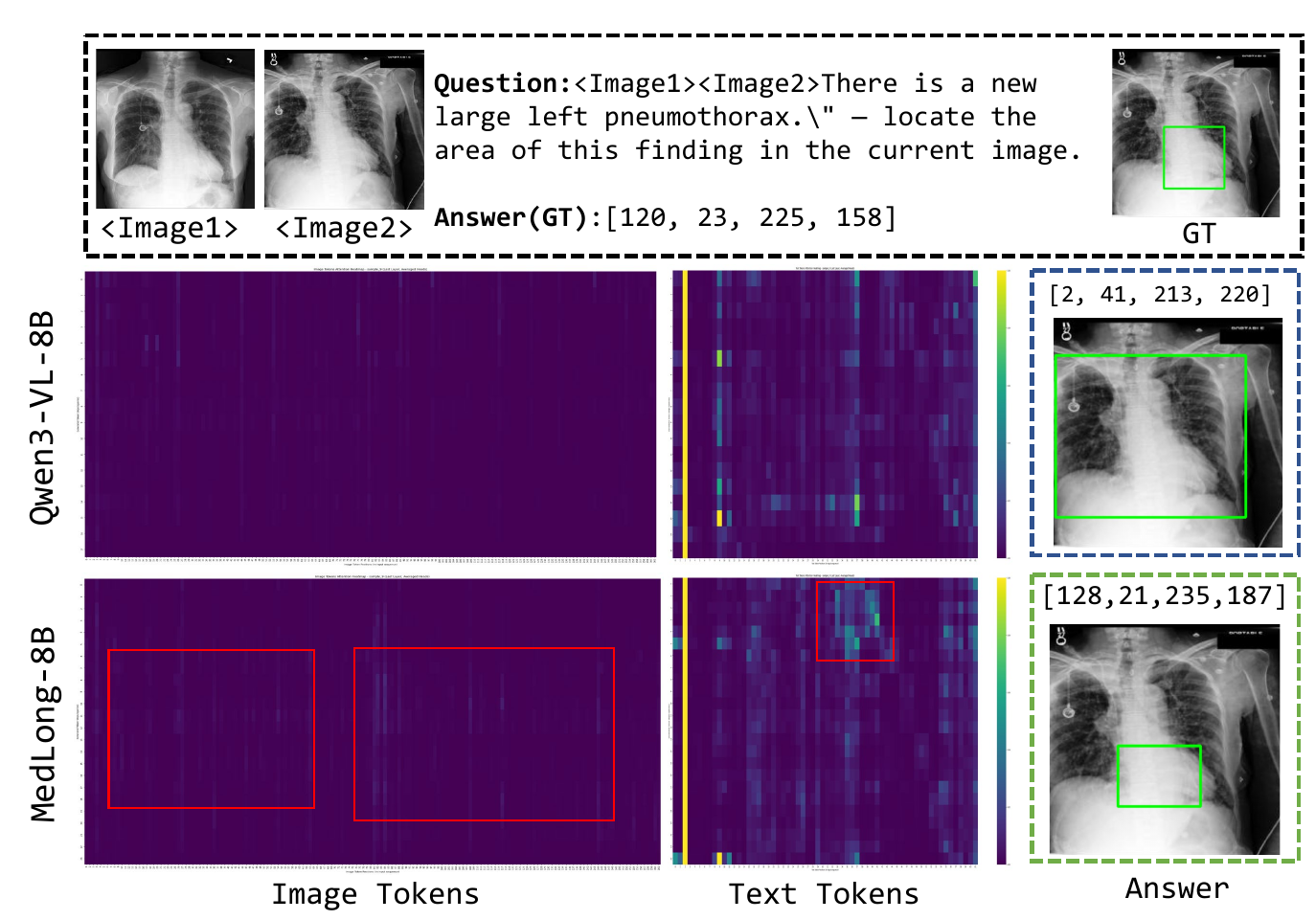}
    \caption{Visualization of attention map during answer generation for task 4 (DRG). The red box indicates the area where attention is enhanced after fine-tuning.}
    \label{fig:attention_4}
\end{figure}

\textbf{Qualitative Analysis}. To further investigate how fine-tuning affects model behavior, we visualize attention maps during answer generation. In~\Cref{fig:attention_4}, the horizontal axis denotes input tokens and the vertical axis denotes generation steps. 
The original model predominantly attends to textual tokens with minimal engagement with visual inputs. 
After fine-tuning, the model first focuses on the semantic intent of the question and progressively increases attention to image regions as generation unfolds. 
This shift indicates a strengthened alignment between visual evidence and linguistic reasoning, leading to more reliable and clinically grounded outputs. 
More visualization results are included in the Appendix~\labelcref{sec:attention_vis}.

\section{Conclusion}
This paper introduces LoMeVQA, a comprehensive multi-task dataset for longitudinal medical image analysis, comprising 206K samples in total. The dataset is structured into two subsets: LoMeVQA-dev for development and training, and LoMeVQA-test for evaluation. To create this dataset, we combine a medical knowledge graph with LLMs to automatically generate longitudinal medical VQA data. Through systematic evaluations on five in-distribution longitudinal VQA tasks together with two out-of-distribution benchmarks, we find that current MLLMs still exhibit significant limitations in longitudinal medical understanding. To bridge this gap, we develop MedLong-4B and MedLong-8B, models specifically trained on the LoMeVQA-dev set. Experimental results show that MedLong-8B substantially outperforms existing MLLMs across all five in-distribution tasks and also generalizes well to the two out-of-distribution benchmarks. We hope that LoMeVQA can facilitate future research on longitudinal medical VQA and advance the development of more capable medical MLLMs.

\bibliographystyle{ACM-Reference-Format}
\bibliography{sample-base}

\clearpage
\onecolumn
\appendix

\section{Dataset Statistics}
\label{sec:dataset_statistic}
~\Cref{tab:dataset_statistic} summarizes the dataset statistics for each task. For every task, we report the total number of samples, as well as the average lengths of images, questions, and answers. These statistics provide a clear view of the dataset’s scale and complexity, which are important considerations for evaluating model performance.
\begin{table}[ht]
  \centering
  \caption{Datasets statistic of LoMeVQA.}
  \label{tab:dataset_statistic}
  \begin{tabular}{l c c c c c c}
    \hline
    Task Type & Size &  Image length &Question Length & Answer Length   & Answer Type\\
    \hline
    Progress Classification & 16925 & 2 & 21.25 & 2.00 & Closed \\
    Progress Description & 36178 & 2 & 12.11 & 21.48 & Open\\
    Progress report generation & 28074 & 3.15 & 16.28 & 40.25 & Open \\
    Differential Region Grounding & 74540 & 2 & 26.19 & 5.00 & Open \\
    Differential Region Description & 51084 & 2 & 18.13 & 14.81 & Open \\
    \hline
  \end{tabular}
\end{table}

\begin{figure}[ht]
    \centering
    \begin{subfigure}[t]{0.45\linewidth}
        \centering
        \includegraphics[width=\linewidth]{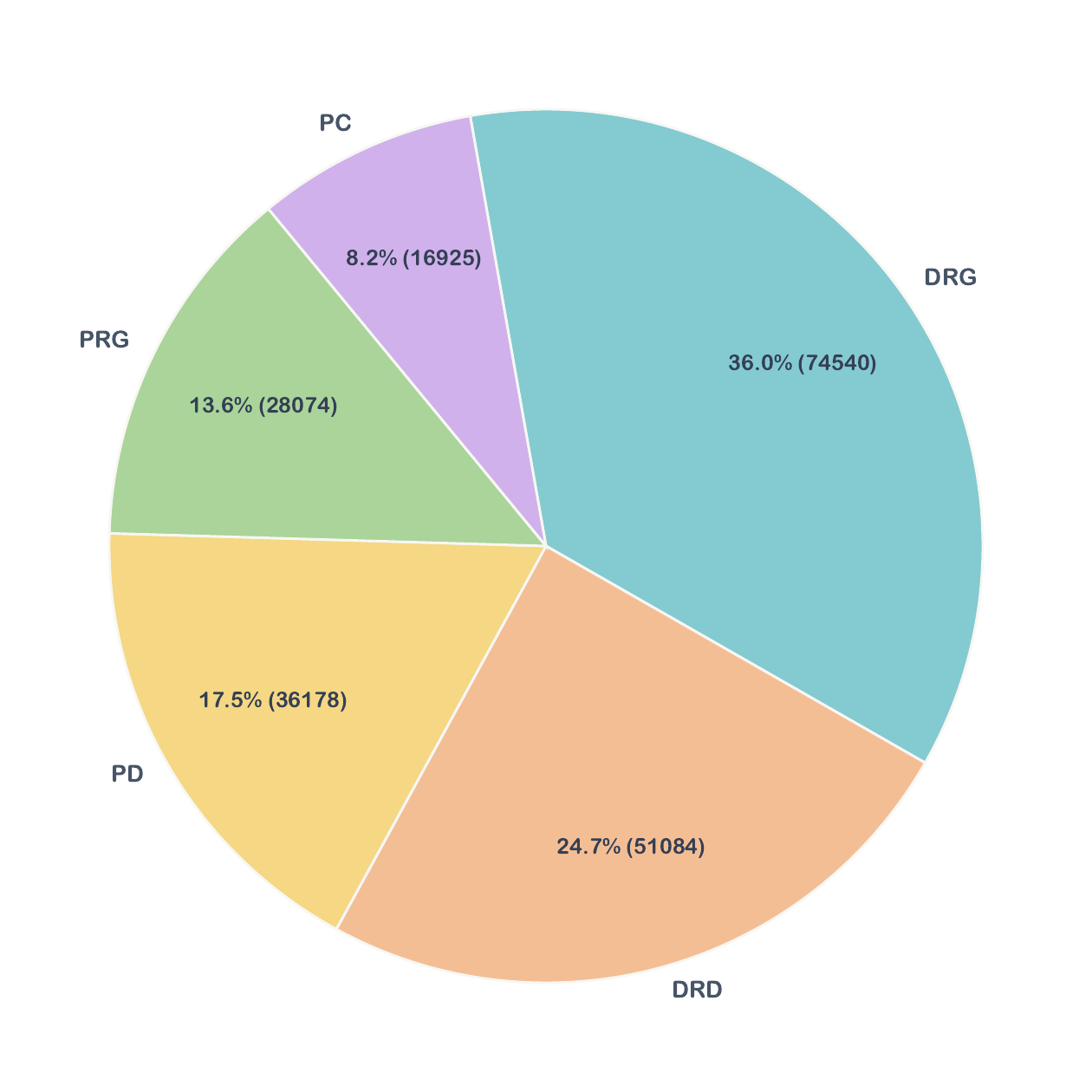}
        \caption{Task sample distribution.}
        \label{fig:task_sample_count_pie}
    \end{subfigure}
    \hfill
    \begin{subfigure}[t]{0.45\linewidth}
        \centering
        \includegraphics[width=\linewidth]{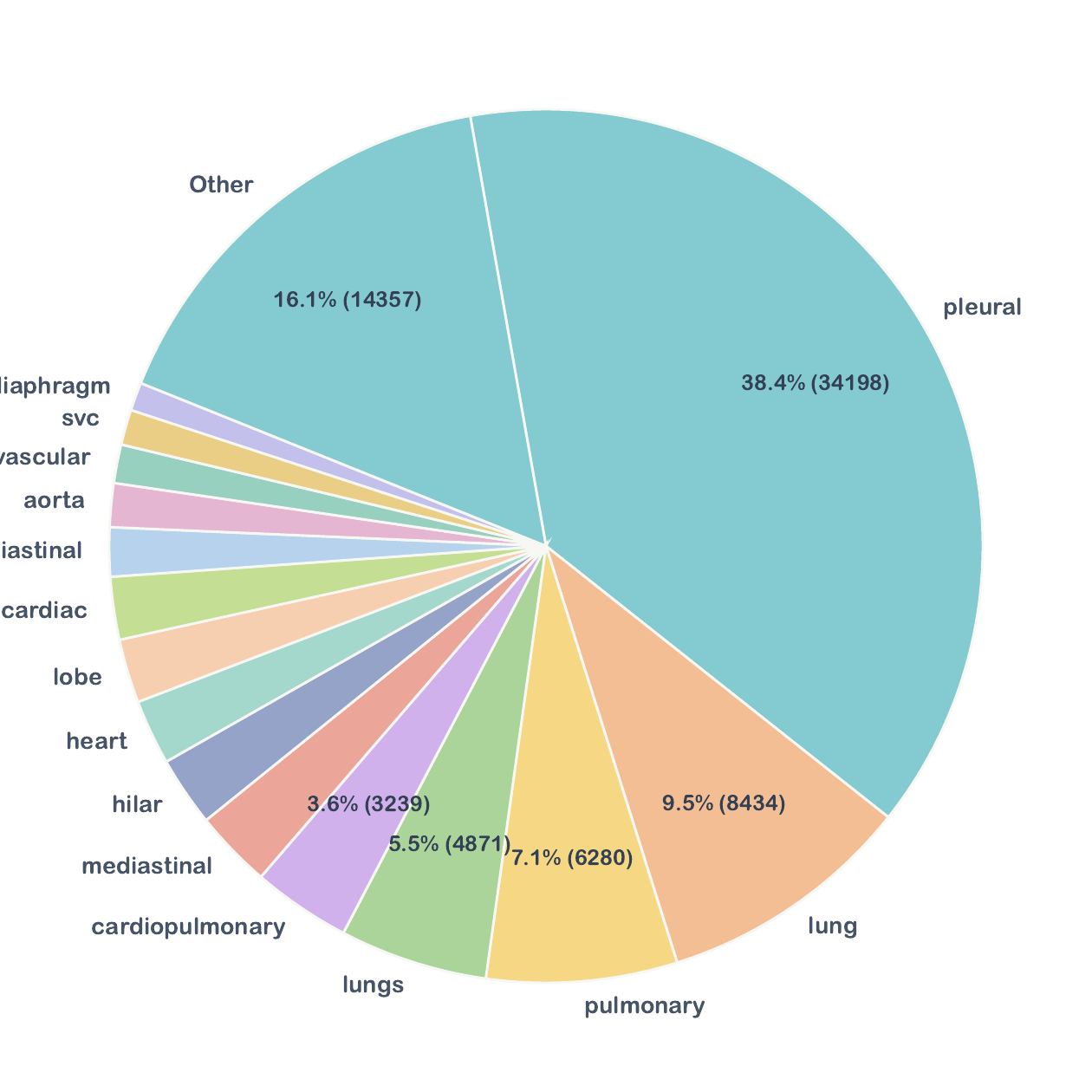}
        \caption{Entity distribution.}
        \label{fig:entity_distribution_pie}
    \end{subfigure}

    \vspace{4pt}
        \begin{subfigure}[t]{0.45\linewidth}
        \centering
        \includegraphics[width=\linewidth]{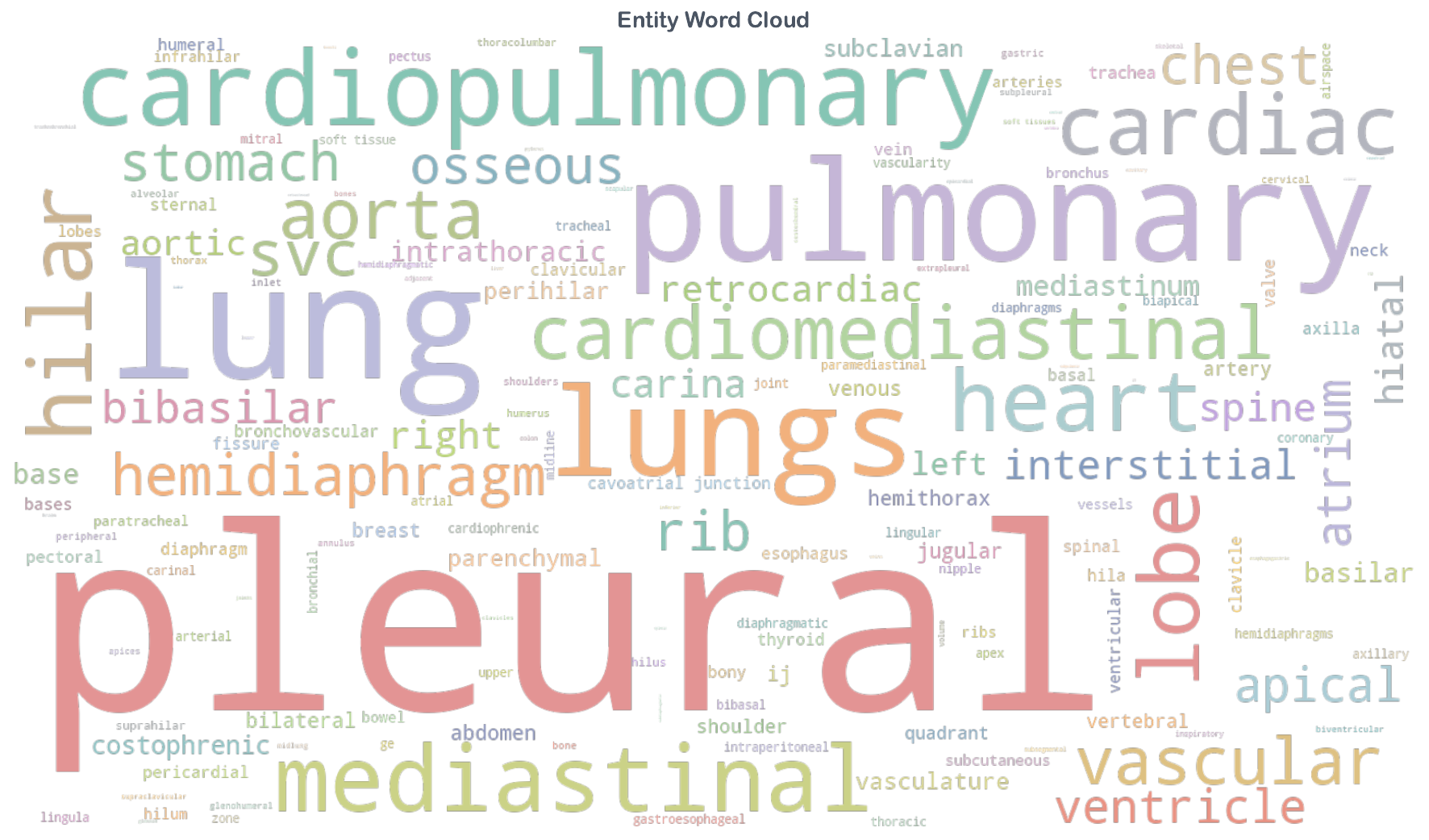}
        \caption{Entity word cloud.}
        \label{fig:entity_wordcloud}
    \end{subfigure}
    \hfill
    \begin{subfigure}[t]{0.45\linewidth}
        \centering
        \includegraphics[width=\linewidth]{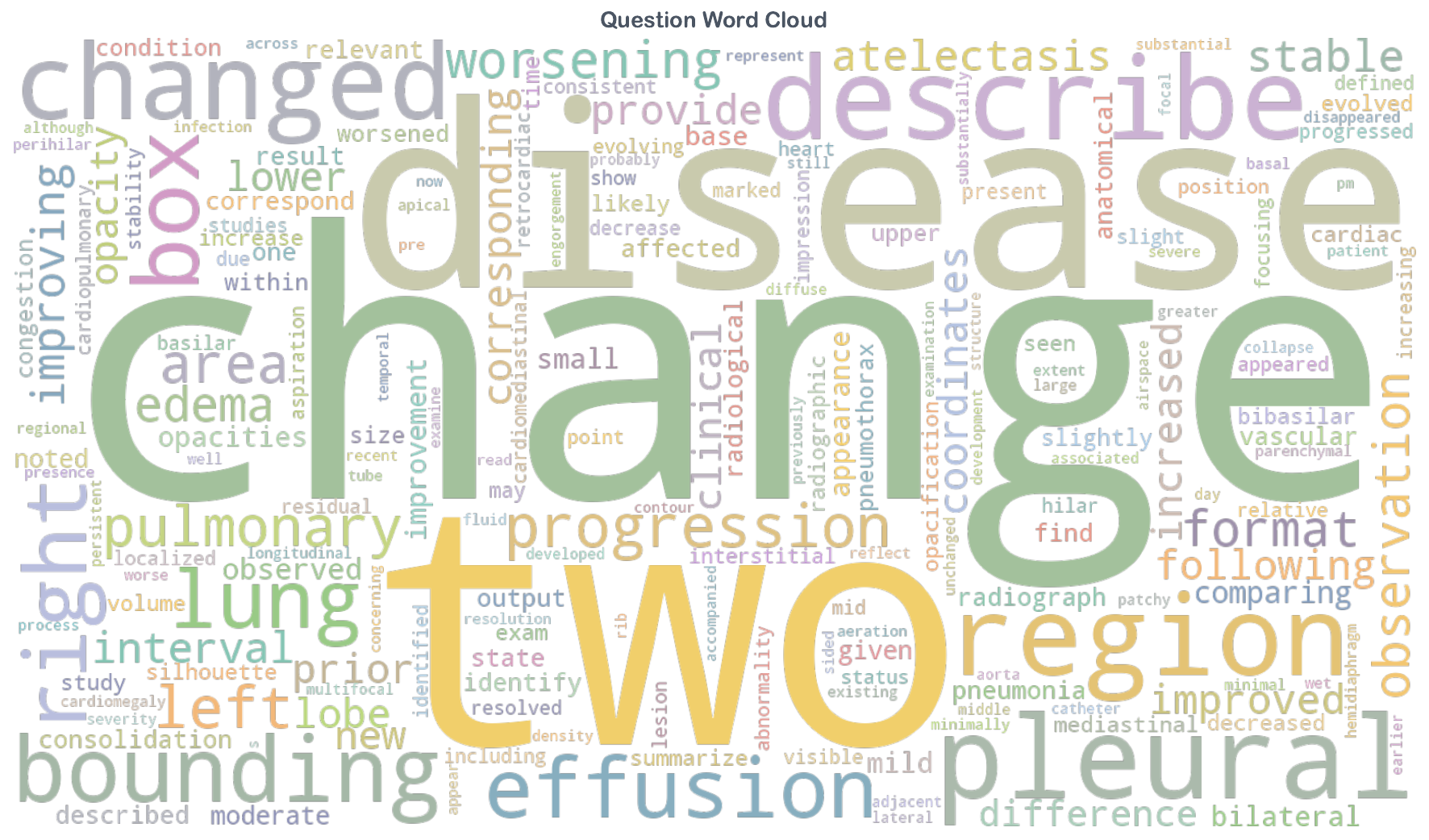}
        \caption{Question word cloud.}
        \label{fig:question_wordcloud}
    \end{subfigure}

    \caption{Dataset statistics and linguistic characteristics. We show the entity distribution, entity word cloud, task sample distribution, and question word cloud.}
    \label{fig:dataset_statistics_overview}
\end{figure}
\clearpage

\section{VQA Generation}
\label{sec:vqa_generation}
\begin{figure}[h]
    \centering
    \includegraphics[width=1\linewidth]{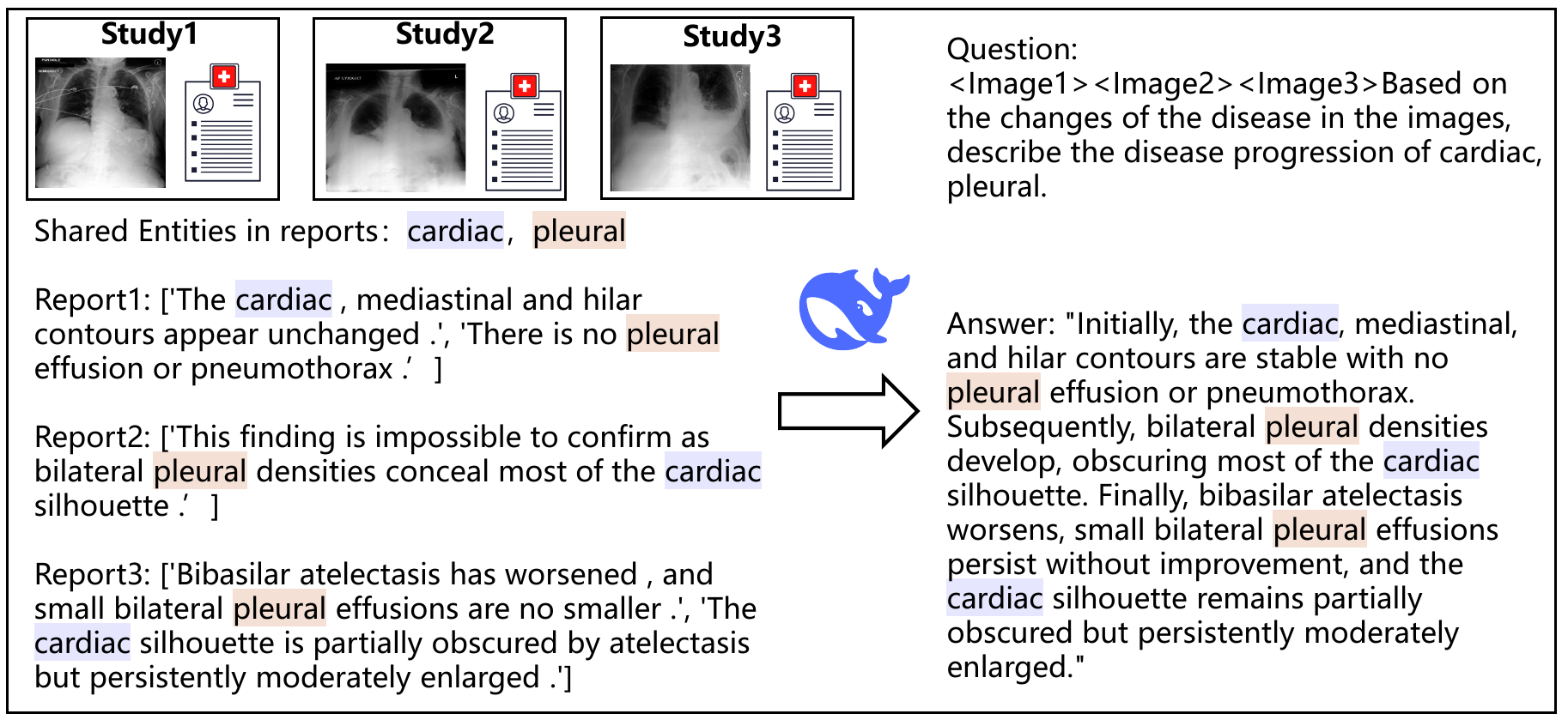}
    \caption{An example for difficulty-controlled VQA generation. We use DeepSeek to generate answers under the setting of K = 3 and N = 2.}
    \label{fig:case_generation}
\end{figure}

\begin{figure}
    \centering
    \includegraphics[width=1.0\linewidth]{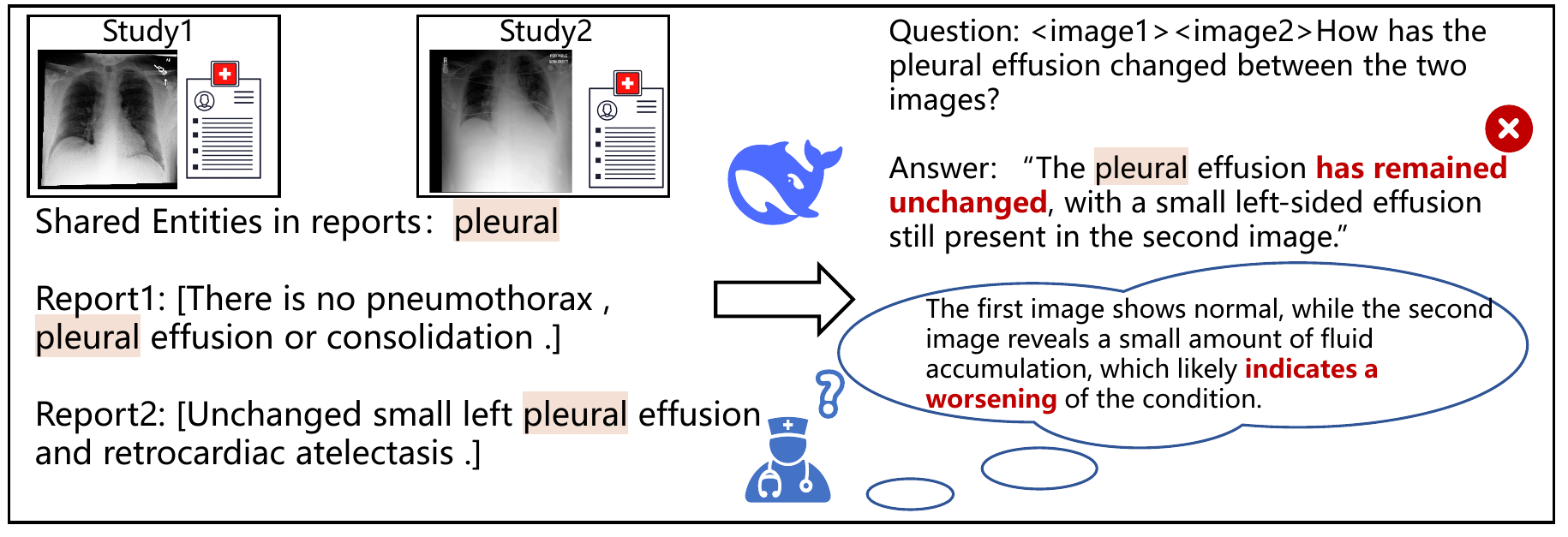}
    \caption{A failure case of during the generation of LLM.}
    \label{fig:failure_generation}
\end{figure}

\Cref{fig:case_generation} shows an example of a difficulty-controlled VQA generation sample. In this example, we construct a sequence of length 3, containing 3 images and 3 corresponding textual descriptions. Each text segment describes 2 entities. We then employ a large language model to generate meaningful questions and answers based on these descriptions, forming the final VQA sample. This process allows us to control the difficulty of the generated VQA pairs while maintaining temporal consistency.
We show a representative failure case of VQA generation in \Cref{fig:failure_generation}. While our pipeline generally produces coherent and clinically meaningful longitudinal VQA samples, some generated examples still contain errors, such as incorrect temporal reasoning, inconsistent entity descriptions, inaccurate region localization, and hallucinated findings. These examples illustrate the main limitations of the current generation pipeline and further motivate the use of quality filtering in dataset construction.

\clearpage
\section{Data Filter}
\label{sec:data_filter}
\begin{figure}[h]
    \centering
    \includegraphics[width=1\linewidth]{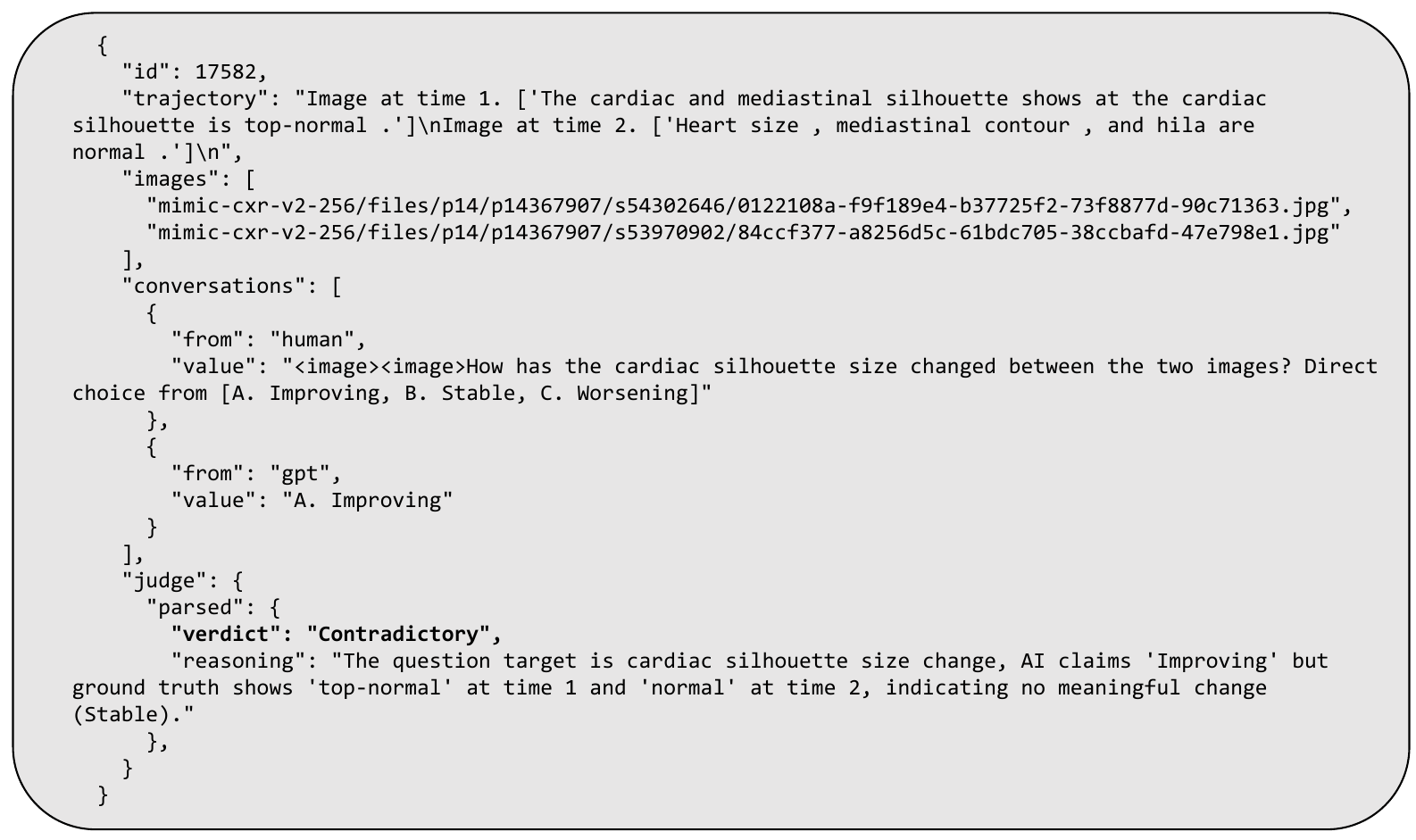}
    \caption{A contradictory sample excluded due to a contradiction between the generated answer and the trajectory in Task 1 (Progress Classification).}
    \label{fig:contradictory}
\end{figure}

\Cref{fig:contradictory} illustrates an example of consistency filtering. In Task 1, the ``conversations" field contains the question-answer pairs generated by the LLM, while the ``judge" field records the LLM's consistency assessment. In this example, the LLM identifies a contradiction between the generated Q\&A and the original data, leading to the removal of this sample from the dataset. This process ensures that only consistent and reliable samples are retained for downstream tasks.

\begin{table}[ht]
\centering
\begin{minipage}[t]{0.48\linewidth}

\centering
\caption{Human--LLM agreement analysis on usability judgments.}
\label{tab:human_llm_check}
\setlength{\tabcolsep}{5pt}
\begin{tabular}{ccccc}
\toprule
\textbf{LLM@Pass} & \textbf{Human@Pass} & \textbf{PC} & \textbf{PD} & \textbf{PRG} \\
\midrule
$\checkmark$ & $\checkmark$ & 203 & 251 & 273 \\
$\checkmark$ & $\times$      & 9   & 16  & 12  \\
$\times$     & $\checkmark$  & 47  & 20  & 9   \\
$\times$     & $\times$      & 41  & 13  & 6   \\
\midrule
\textbf{Overall} &  & 300 & 300 & 300 \\
\textbf{Gwet's AC1} &  & 0.71 & 0.85 & 0.92 \\
\bottomrule
\end{tabular}
\end{minipage}
\hfill
\begin{minipage}[t]{0.48\linewidth}
\centering
\caption{Accuracy of three clinical experts in manually reviewing LoMeVQA-test.}
\label{tab:human_check}
\setlength{\tabcolsep}{4pt}
\begin{tabular}{lcccc}
\toprule
\textbf{Task} & \textbf{Junior} & \textbf{Intermediate} & \textbf{Senior} & \textbf{Human Avg.} \\
\midrule
PC  & 94\% & 98\% & 94\% & 95.33\% \\
PD  & 96\% & 92\% & 92\% & 93.33\% \\
PRG & 96\% & 94\% & 92\% & 94.00\% \\
DRG & 96\% & 92\% & 96\% & 94.67\% \\
DRD & 94\% & 90\% & 93\% & 92.33\% \\
\midrule
\textbf{Average} & 95.20\% & 93.20\% & 93.40\% & 93.93\% \\
\bottomrule
\end{tabular}
\end{minipage}
\end{table}

\clearpage

\section{Case Studies}
\label{sec:case_studies}
In addition to the analysis of Task 2 presented in \Cref{sec:experiments}, we provide case studies for the remaining four tasks. These qualitative examples illustrate the performance of different MLLMs on the LoMeVQA-test set, highlighting their strengths and limitations across various tasks.

\begin{figure}[h]
    \centering
    \includegraphics[width=1\linewidth]{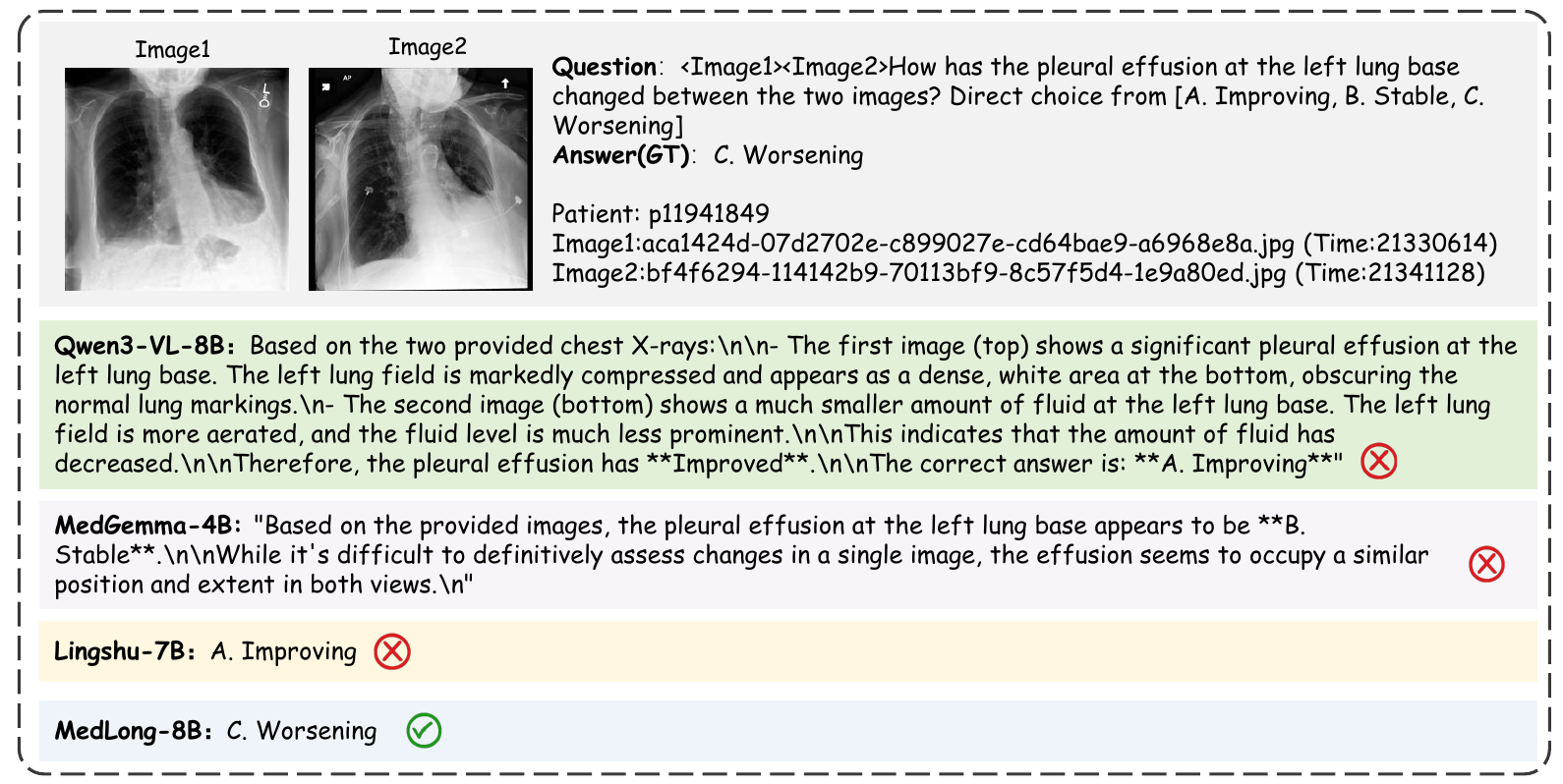}
    \caption{A case study for task 1 (Progress Classification).}
\end{figure}

\begin{figure}
    \centering
    \includegraphics[width=1\linewidth]{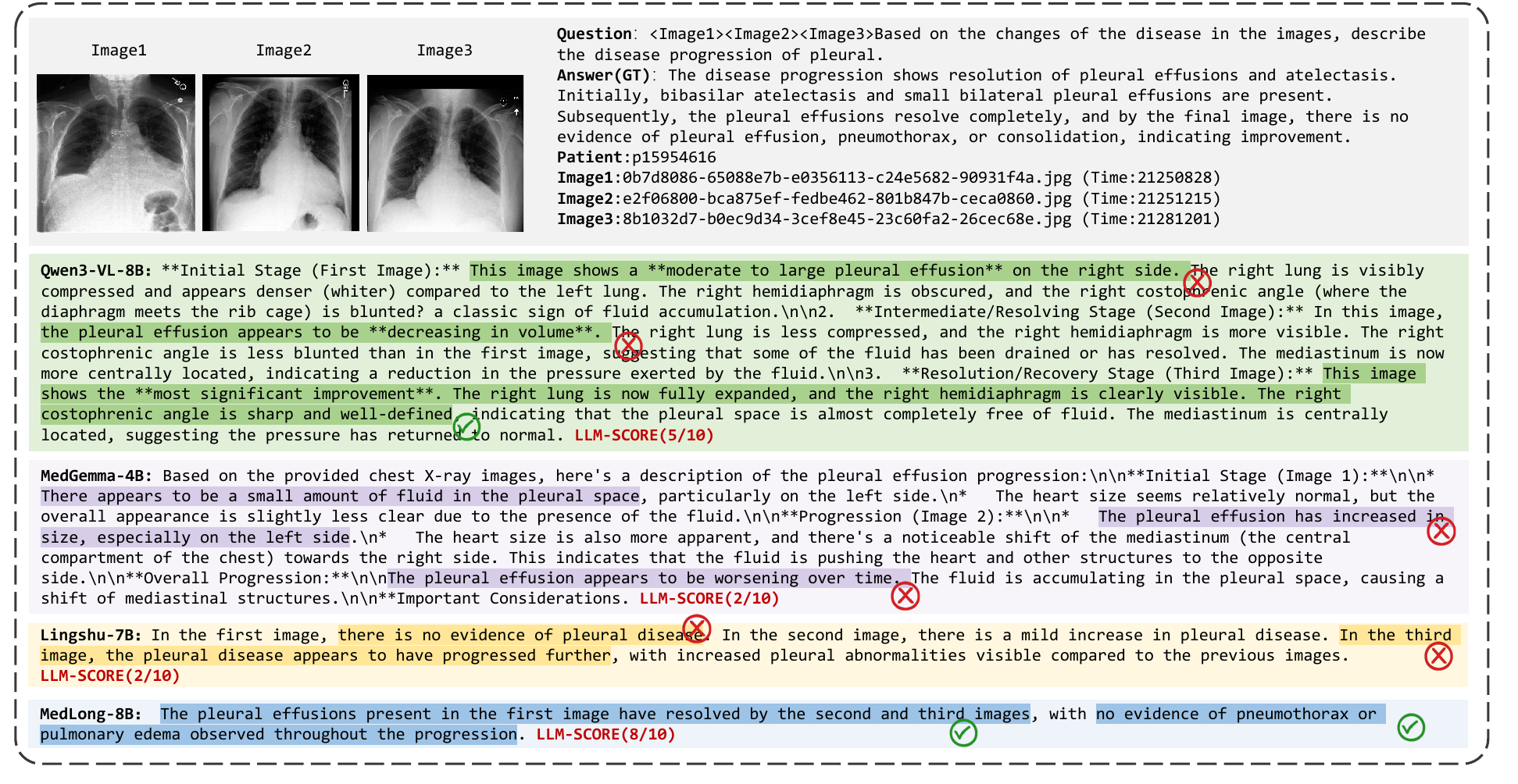}
    \caption{A case study for task 3 (Progress Report Generation).}
\end{figure}

\begin{figure}
    \centering
    \includegraphics[width=1\linewidth]{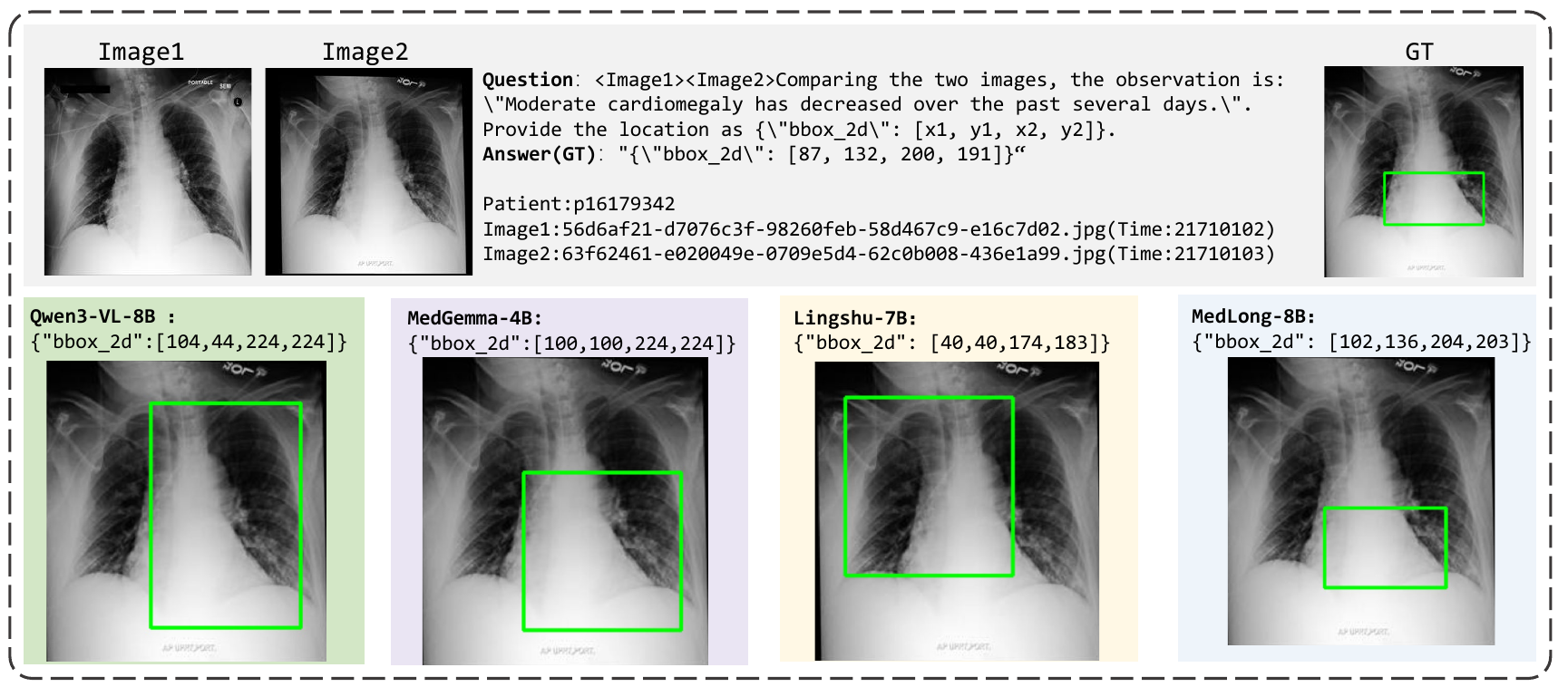}
    \caption{A case study for task 4 (Differential Region Grounding).}
\end{figure}

\begin{figure}
    \centering
    \includegraphics[width=1\linewidth]{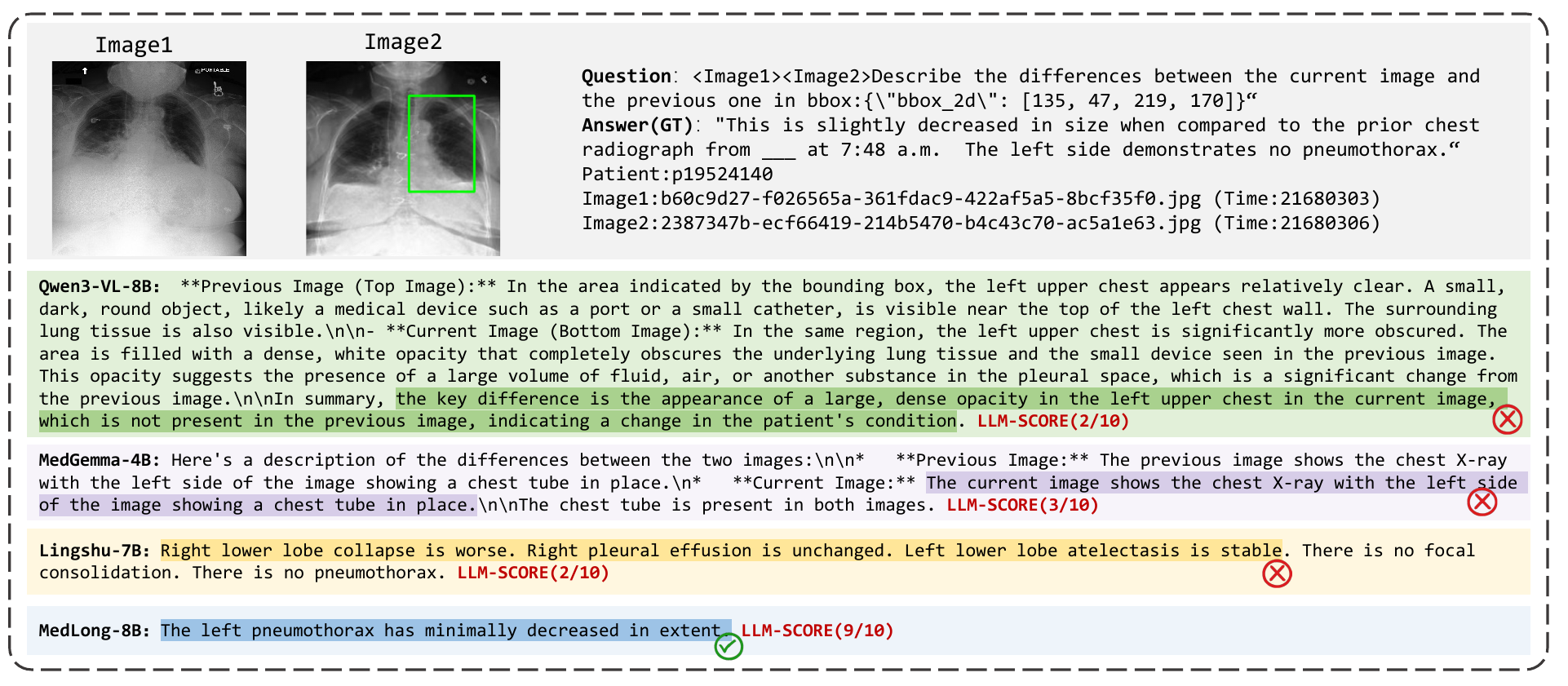}
    \caption{A case study for task 5 (Differential Region Description).}
\end{figure}

\section{LLM-As-A-Judge}
\label{llm-as-a-judge}
\begin{figure}
    \centering
    \includegraphics[width=1\linewidth]{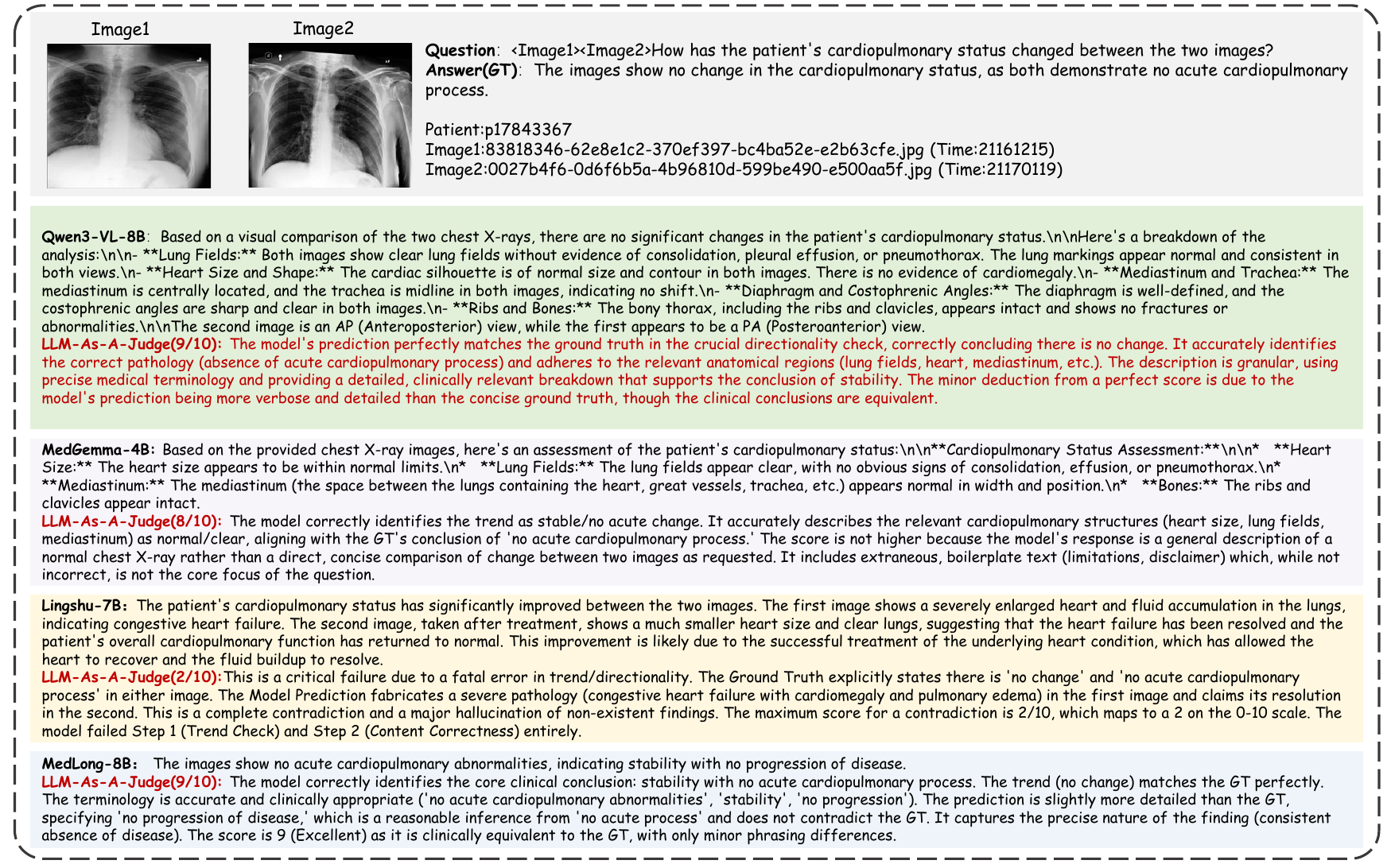}
    \caption{Illustration of LLM-as-a-Judge for evaluating open-ended medical responses. }
    \label{fig:LLM-As-A-Judge}
\end{figure}

\Cref{fig:LLM-As-A-Judge} shows an example of the LLM scoring reasoning process. To better evaluate open-ended responses, we adapt an LLM-as-a-Judge protocol due to the limitations of conventional NLG metrics such as ROUGE and METEOR. Since these lexical-overlap based metrics tend to favor template-like outputs and are less sensitive to clinical correctness and temporal consistency, they are insufficient for evaluating longitudinal medical responses. We therefore use an external judge model via the DeepSeek API for reference-aware scoring. For each sample, the judge model takes the question, ground-truth answer, and model prediction as input, and returns a structured JSON output containing a scalar score and textual rationale. The scoring prompt is tailored to longitudinal medical evaluation, focusing on: (1) trend and directionality consistency between the prediction and reference, (2) content correctness and hallucination control, including pathology and region consistency, and (3) descriptive granularity and terminology accuracy. The judge assigns an integer score from 0 to 10, with contradictory progression trends treated as critical errors and heavily penalized.

\clearpage
\section{Attention Visualization}
In addition to the examples presented in \Cref{sec:experiments}, we provide attention maps for the remaining tasks. These visualizations illustrate how the model attends to different regions of the input images and text across various tasks.
\label{sec:attention_vis}
\begin{figure}[h]
    \centering
    \includegraphics[width=1\linewidth,height=8cm,keepaspectratio]{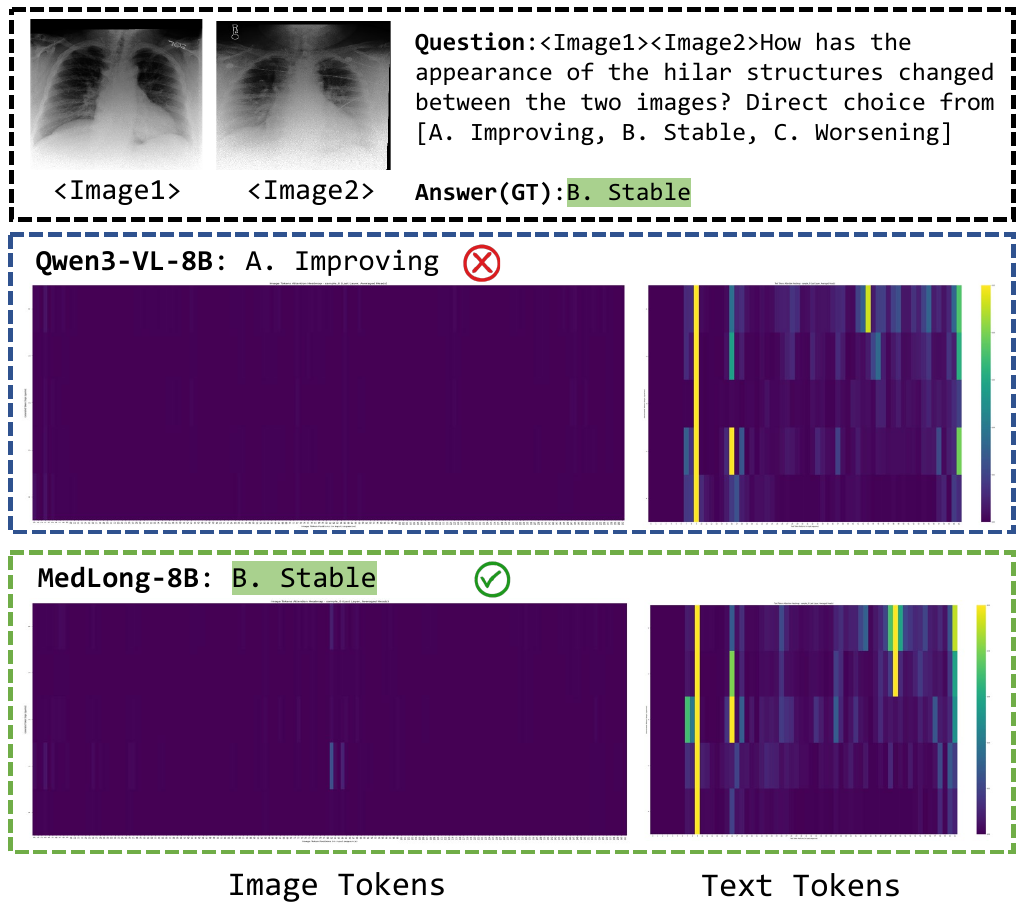}
    \caption{Attention map for task 1 (Progress Classification).}
\end{figure}

\begin{figure}[ht]
    \centering
    \includegraphics[width=1\linewidth,height=9cm,keepaspectratio]{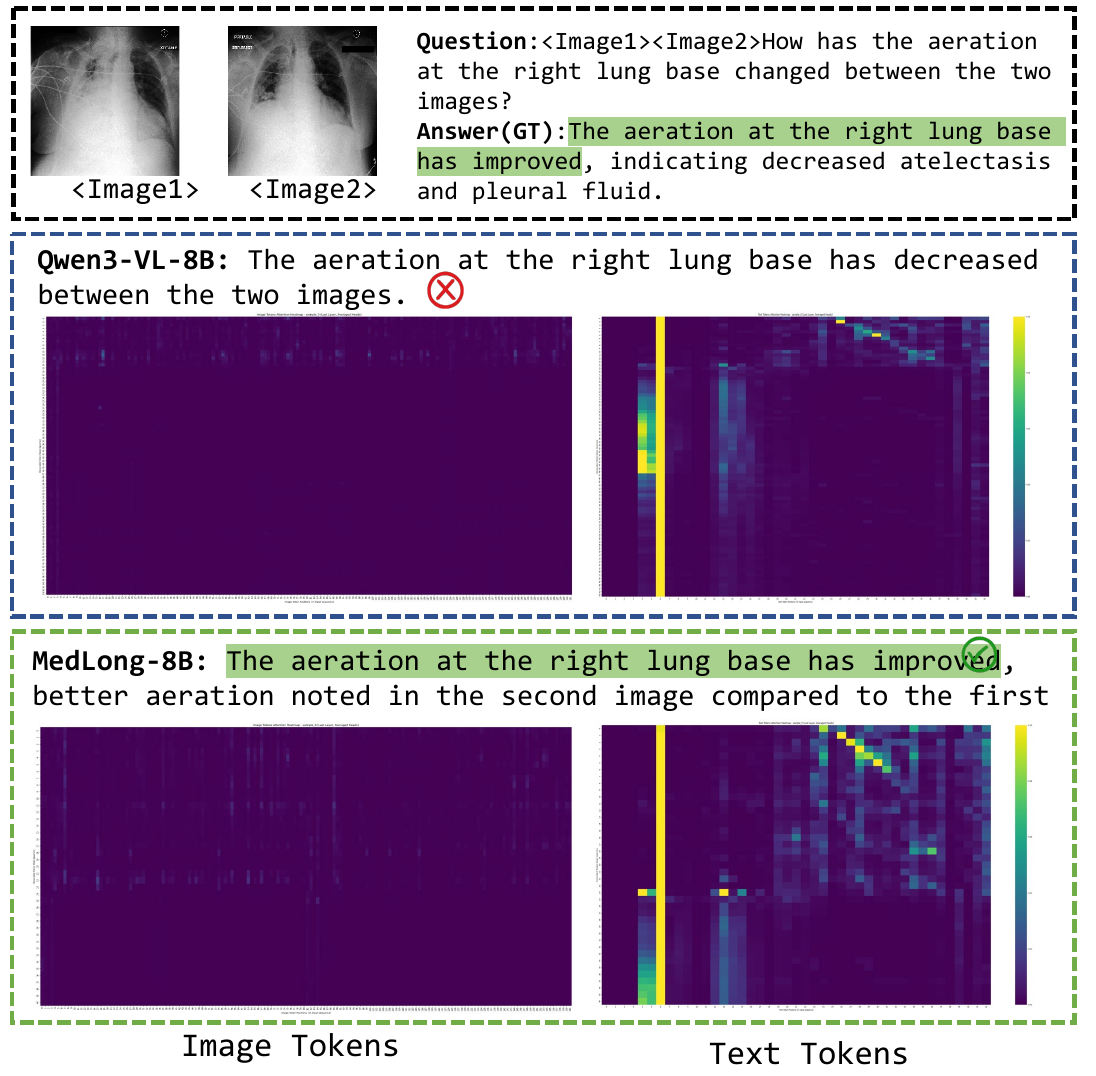}
    \caption{Attention map for task 2 (Progress Description).}
\end{figure}

\begin{figure}[ht]
    \centering
    \includegraphics[width=0.9\linewidth]{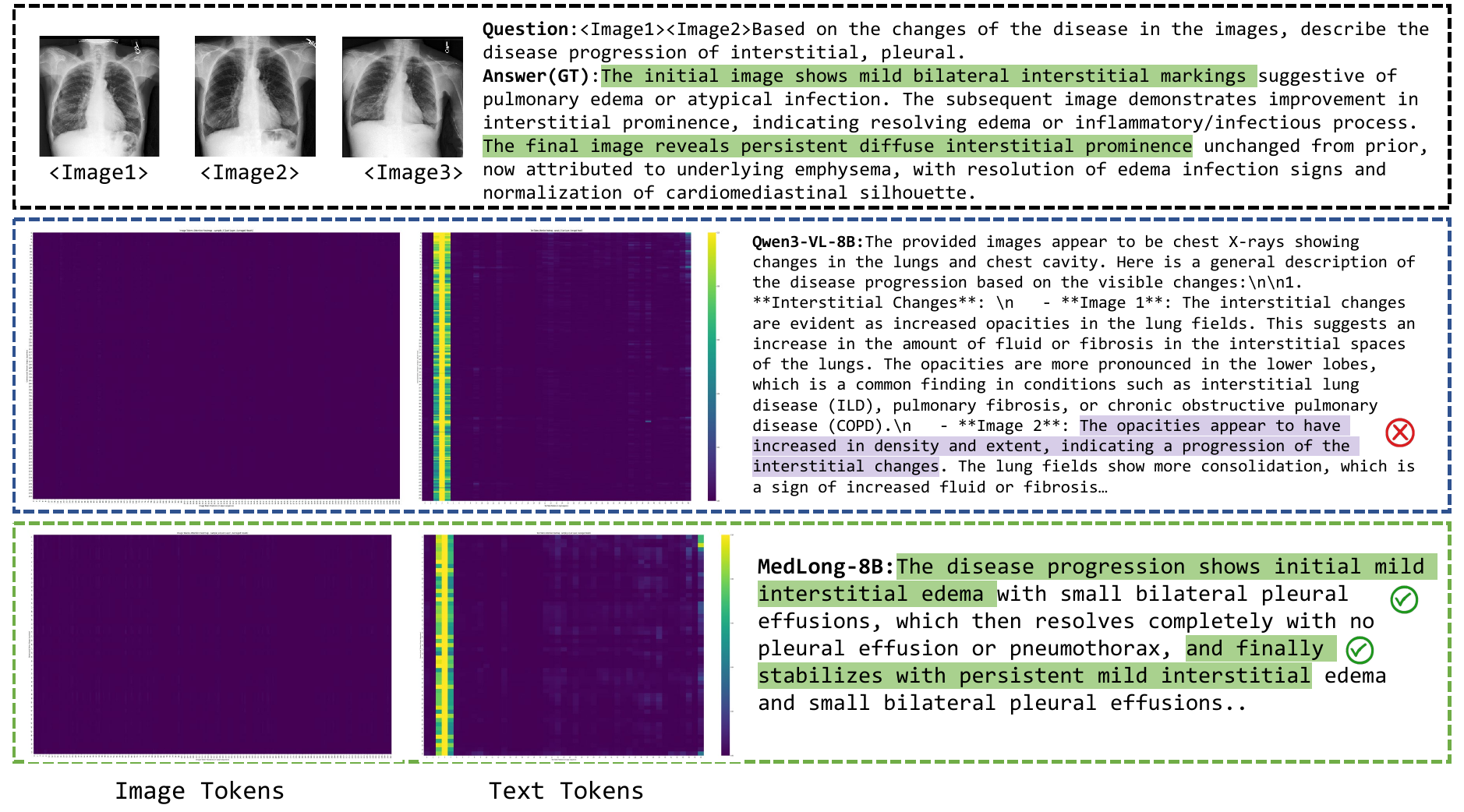}
    \caption{Attention map for task 3 (Progress Report Generation).}
\end{figure}

\begin{figure}
    \centering
    \includegraphics[width=1\linewidth,height=11cm,keepaspectratio]{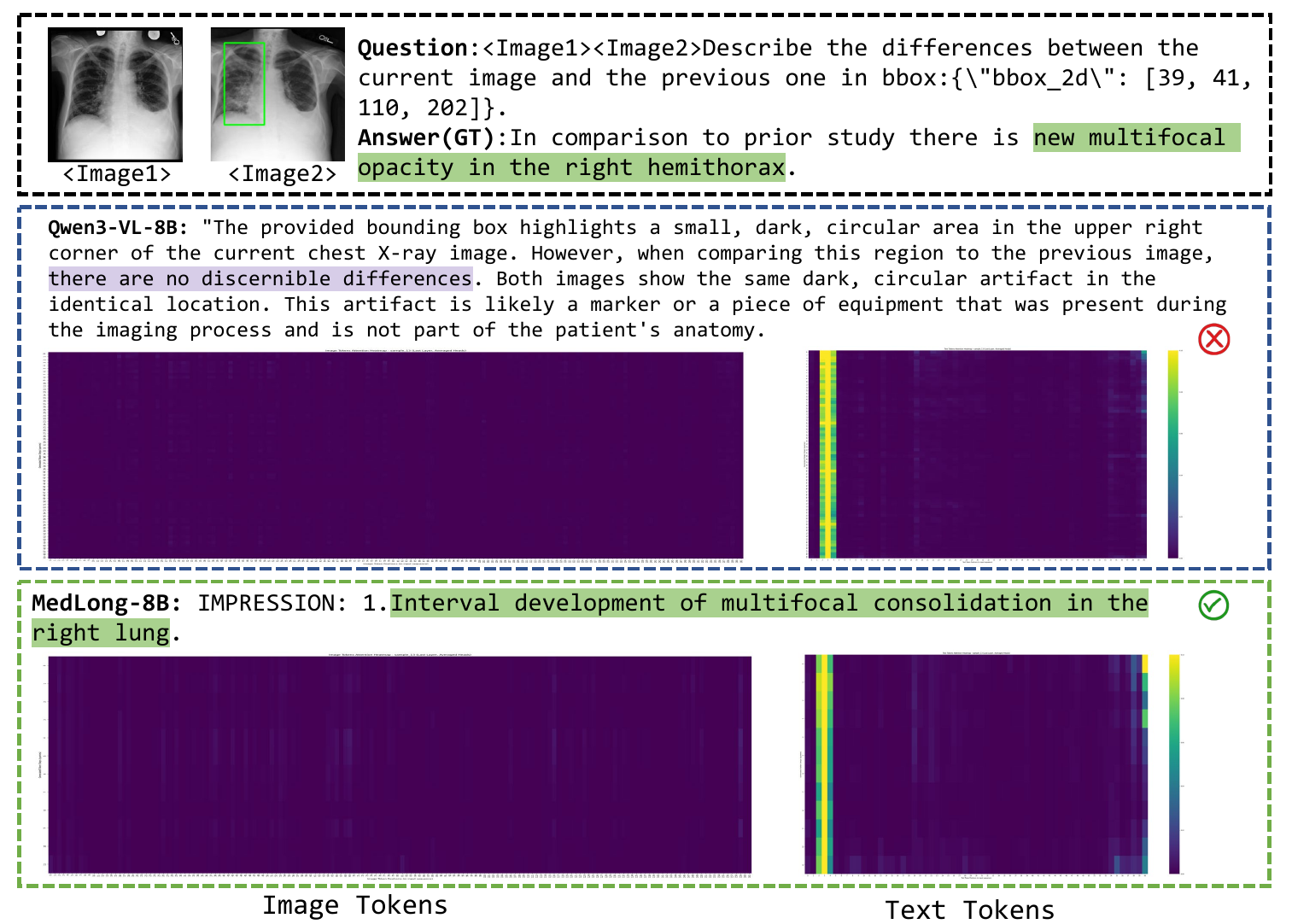}
    \caption{Attention map for task 5 (Differential Region Description).}
\end{figure}

\clearpage
\section{Prompt Details}
\label{sec:prompt_details}
In this section, we provide the prompts used when interacting with the LLM. 
\begin{figure}[h]
    \centering
    \includegraphics[width=1.0\linewidth]{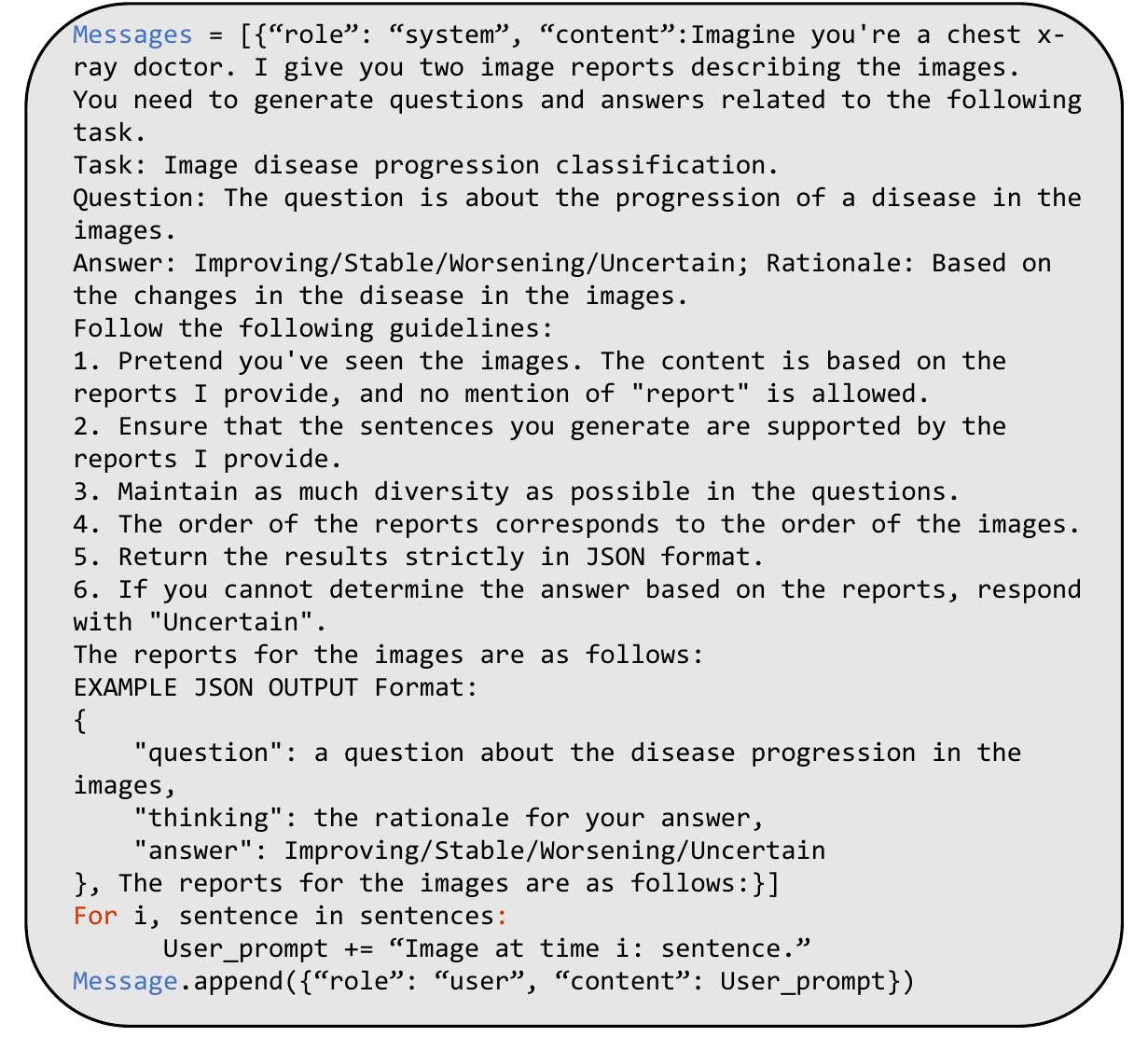}
    \caption{Prompt for VQA pairs generation of task 1 (Progress Classification).}
\end{figure}

\begin{figure}
    \centering
    \includegraphics[width=1.0\linewidth]{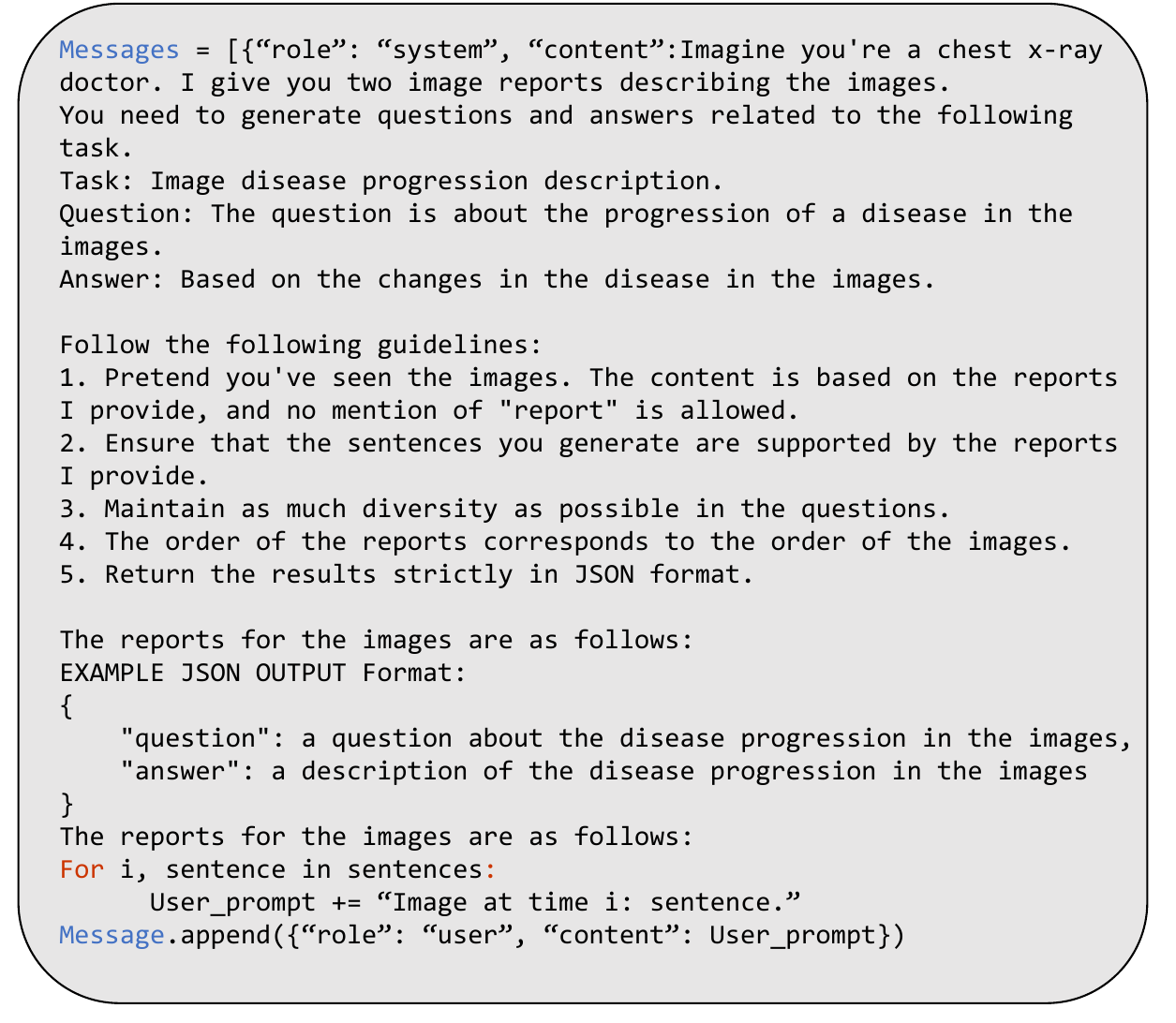}
    \caption{Prompt for VQA pairs generation of task 2 (Progress Description).}
\end{figure}

\begin{figure}
    \centering
    \includegraphics[width=1.0\linewidth]{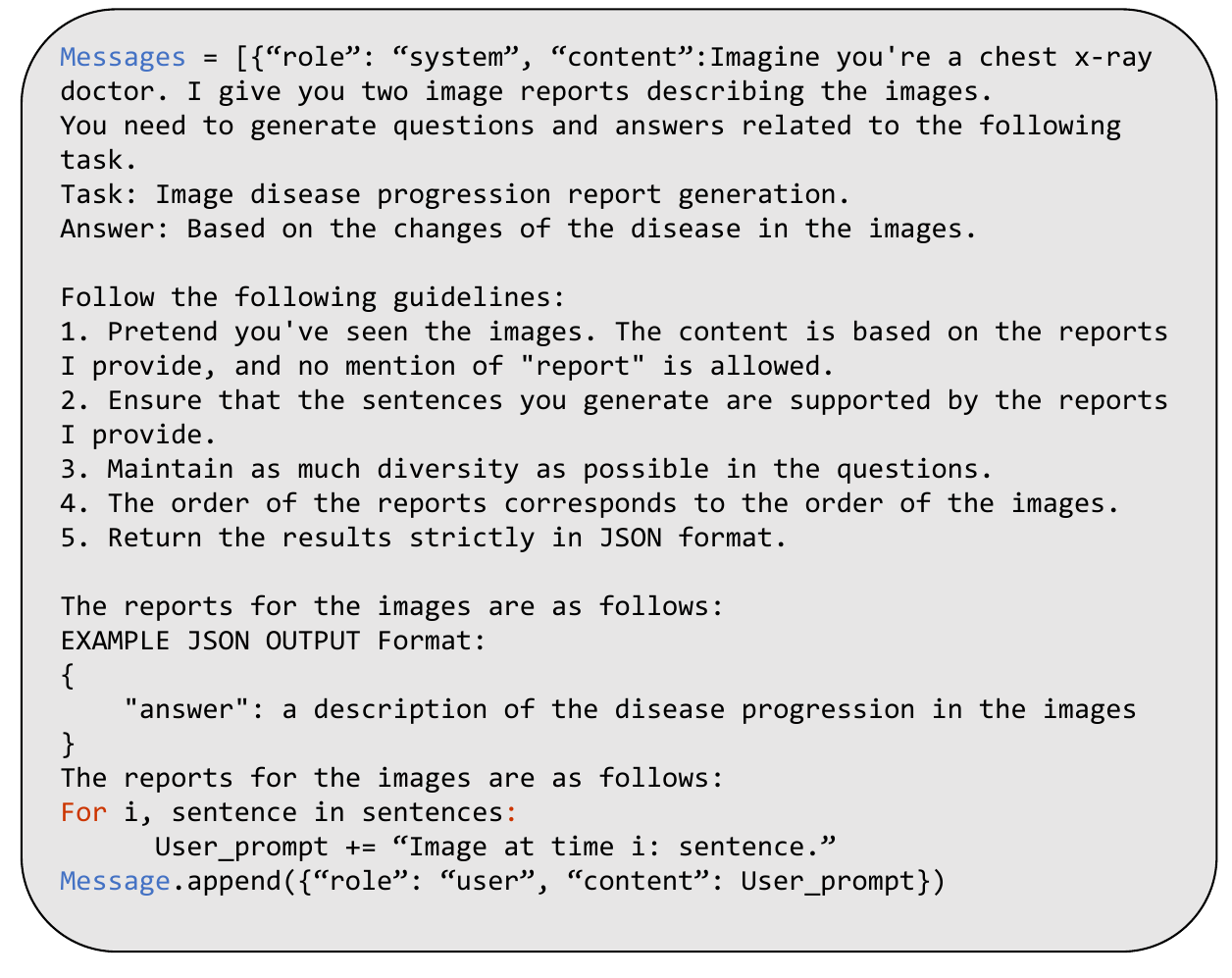}
    \caption{Prompt for VQA pairs generation of task 3 (Progress Report Generation ).}
\end{figure}

\begin{figure}
    \centering
    \includegraphics[width=1.0\linewidth]{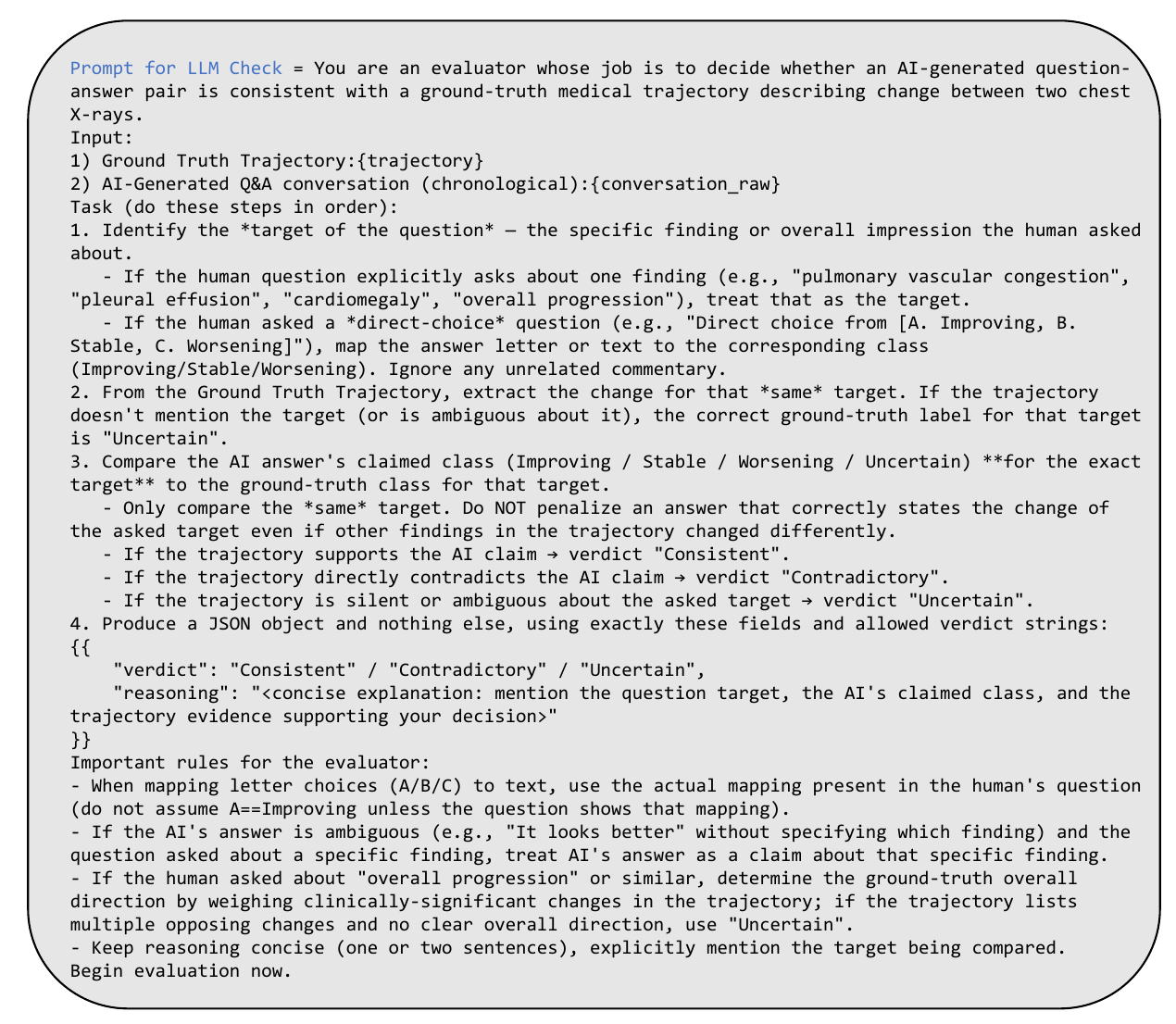}
    \caption{Prompt for the consistency check using LLM.}
    \label{fig:placeholder}
\end{figure}

\begin{figure}
    \centering
    \includegraphics[width=1.0\linewidth]{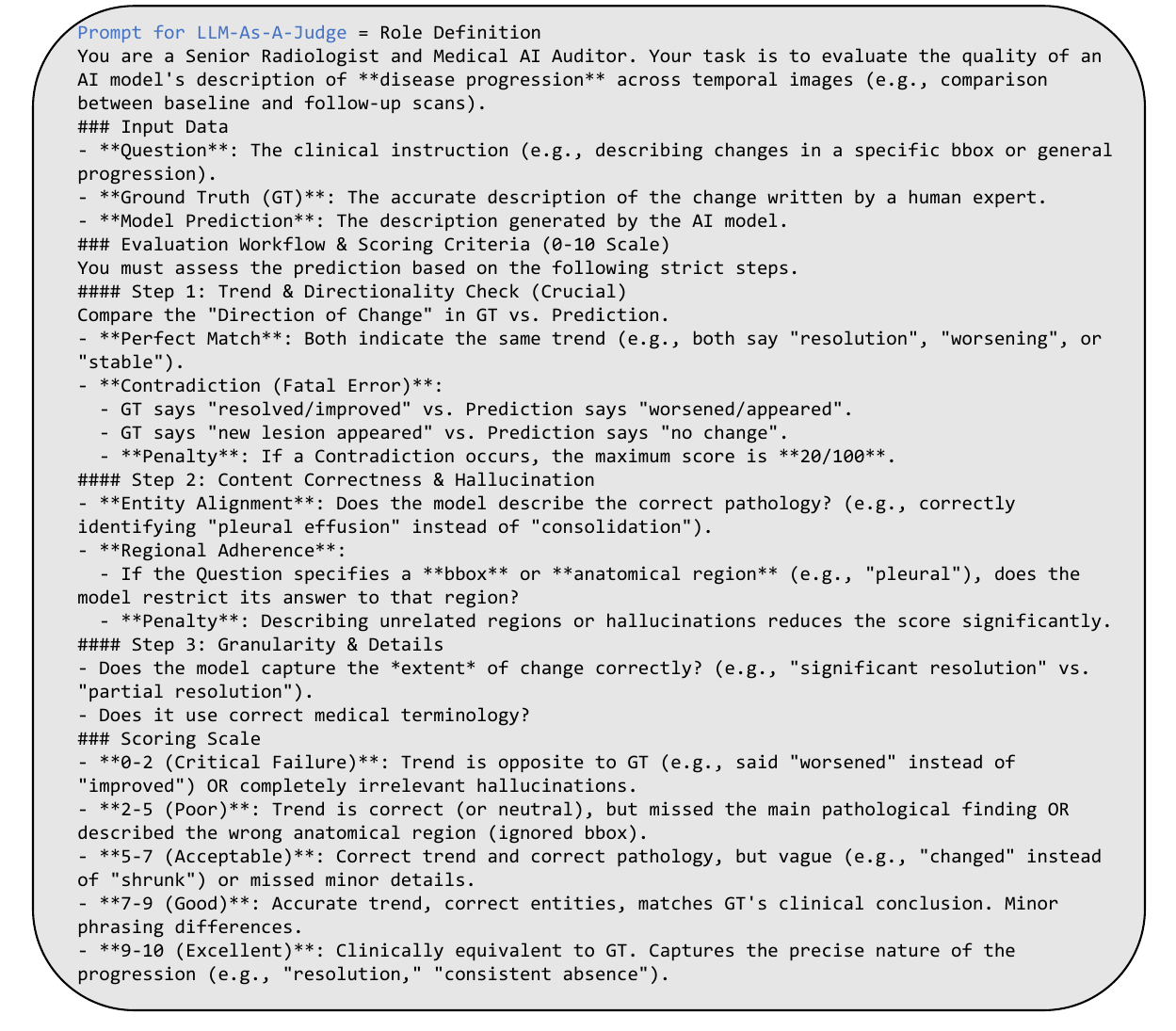}
    \caption{The prompt for LLM-As-A-Judge}
    \label{fig:prompt_llm_as_a_judge}
\end{figure}

\end{document}